\documentclass[runningheads]{llncs}

\usepackage[review,year=2026,ID=11333]{eccv}
\usepackage{eccvabbrv}

\usepackage{graphicx}
\usepackage{booktabs}

\usepackage[accsupp]{axessibility}  % Improves PDF readability for those with disabilities.

\usepackage{hyperref}

\usepackage{orcidlink}

\usepackage{graphicx}
\usepackage{siunitx}
\usepackage{multirow}
\usepackage{algorithm}
\usepackage{algpseudocode}
\usepackage{amsmath}
\usepackage{arydshln}

\begin{document}

\title{Tuning Real-World Image Restoration at Inference: A Test-Time Scaling Paradigm for Flow Matching Models} 


\author{Purui Bai\inst{1,2}\orcidlink{0009-0003-7080-2180} \and
Junxian Duan\inst{1}\orcidlink{0000-0002-0218-6924} \and
Pin Wang\inst{1}\orcidlink{0009-0008-8697-9569} \and
Jinhua Hao\inst{2}\orcidlink{0000-0003-4571-2063} \and
Ming Sun\inst{2}\orcidlink{0000-0002-8625-5199} \and 
Chao Zhou\inst{2}\orcidlink{0000-0002-8625-5199} \and
Huaibo Huang\inst{1}\orcidlink{0000-0001-5866-2283}}


\institute{MAIS \& NLPR, Institute of Automation, Chinese Academy of Sciences, China \and
Kuaishou Tech, China}

\maketitle

\begin{abstract}
Although diffusion-based real-world image restoration (Real-IR) has achieved remarkable progress, efficiently leveraging ultra-large-scale pre-trained text-to-image (T2I) models and fully exploiting their potential remain significant challenges. To address this issue, we propose \textbf{ResFlow-Tuner}, an image restoration framework based on the state-of-the-art flow matching model, FLUX.1-dev, which integrates unified multi-modal fusion (UMMF) with test-time scaling (TTS) to achieve unprecedented restoration performance. Our approach fully leverages the advantages of the Multi-Modal Diffusion Transformer (MM-DiT) architecture by encoding multi-modal conditions into a unified sequence that guides the synthesis of high-quality images. Furthermore, we introduce a training-free test-time scaling paradigm tailored for image restoration. During inference, this technique dynamically steers the denoising direction through feedback from a reward model (RM), thereby achieving significant performance gains with controllable computational overhead. Extensive experiments demonstrate that our method achieves state-of-the-art performance across multiple standard benchmarks. This work not only validates the powerful capabilities of the flow matching model in low-level vision tasks but, more importantly, proposes a novel and efficient inference-time scaling paradigm suitable for large pre-trained models.
  \keywords{Image Restoration \and Test-Time Scaling \and Flow Matching Model}
\end{abstract}

\begin{figure}[tb]
    \centering
    \includegraphics[width=\textwidth]{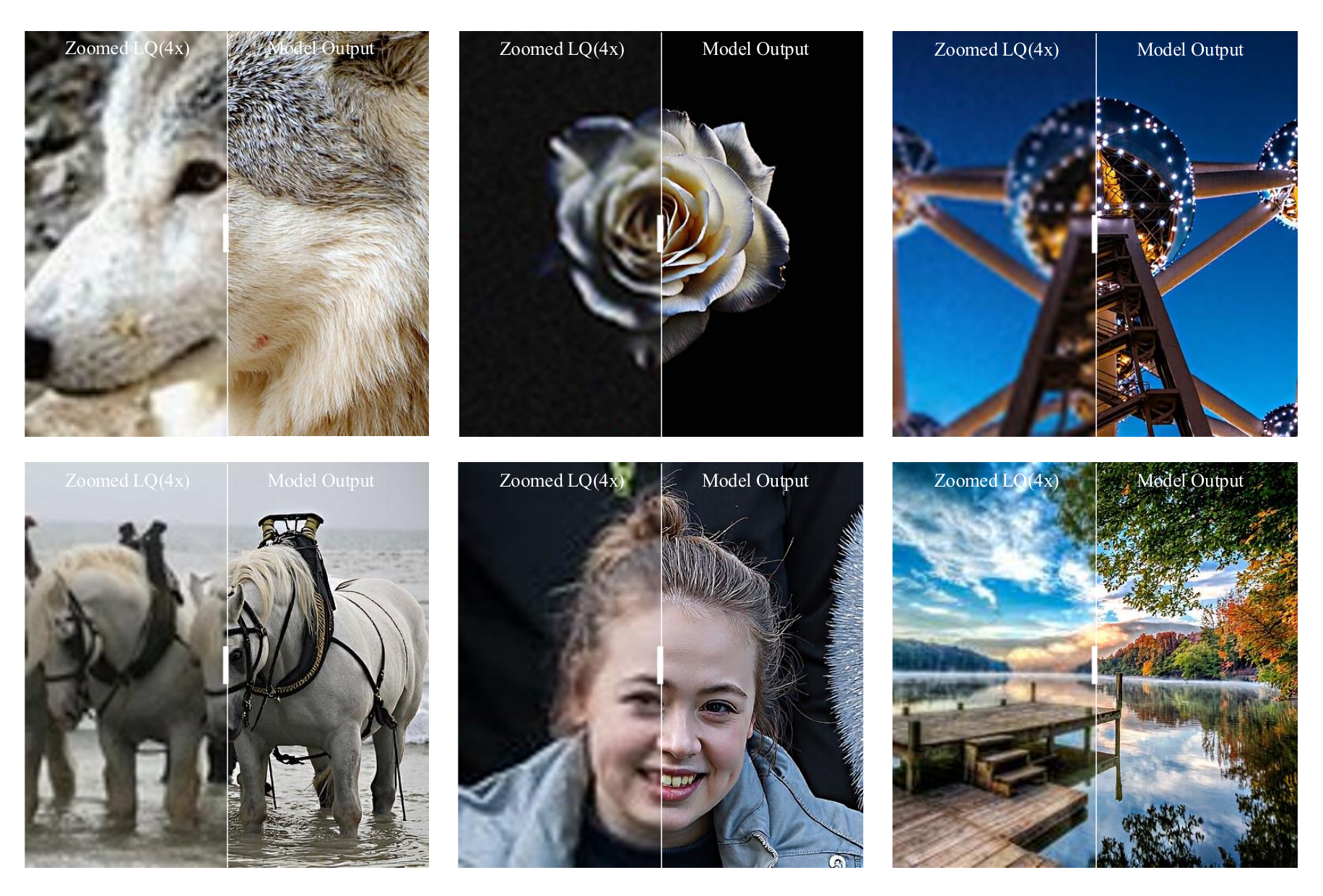}
    \caption{\textbf{ResFlow-Tuner} delivers superior performance on both synthetic (the first row) and real-world (the second row) benchmarks, excelling in terms of perceptual quality and objective image quality assessment.}
    \label{fig:1}
\end{figure}

\section{Introduction}
\label{sec:intro}

Image restoration (IR), a vital field in computer vision, targets transforming degraded low-quality (LQ) images into high-quality (HQ) counterparts. While IR has achieved significant advancements under predefined conditions, such as super-resolution \cite{chen2024ntire, ren2025tenth} and denoising \cite{sun2025tenth} tasks, real-world IR remains a formidable challenge due to the diversity and complexity of degradation types.

The field of image restoration has been revolutionized by generative models. Numerous studies have shown that, in terms of restored image quality, SD-based methods outperform GAN-based ones. While diffusion models, which learn to reverse a stochastic degradation process, have set impressive benchmarks, a new paradigm is emerging: flow-based generative models \cite{flux2024}, which leverage Ordinary Differential Equations (ODEs) to learn a deterministic and often more efficient mapping from noise to data. Previous research \cite{lipman2022flow, liu2022flow, kim2025inference}has shown that this ODE formulation captures a more coherent trajectory in the latent space, leading to superior sample quality and finer granularity in generated details compared to their Stochastic Differential Equation (SDE)-based diffusion counterparts.

However, harnessing such powerful flow-based foundations for image restoration (IR) presents unique challenges. First, the scale of models like FLUX makes full fine-tuning computationally intractable. While parameter-efficient methods (e.g., LoRA \cite{hu2022lora}) are necessary, they alone may not fully exploit the model's capabilities. Second, mainstream conditional mechanisms for IR typically employ a relatively straightforward method \cite{lin2024diffbir, wang2024exploiting}of direct conditional injection, limiting the utilization of priors in pre-trained models. In this paper, we introduce \textbf{ResFlow-Tuner}, a novel framework not only adapts the flow model for IR but also introduces a pioneering test-time scaling (TTS) strategy specifically designed for its ODE dynamics. Our framework features a Unified Multi-Modal Fusion (UMMF) mechanism \cite{esser2024scaling}. This design enables all semantic tags to interact directly within the multimodal attention mechanism, thereby ensuring highly faithful reconstruction that is accurate to both the pixel-level input and the high-level semantic guidance. Besides, we redesign the training-free TTS by introducing perturbations to intermediate states along the denoising ODE trajectory. The quality of these perturbed states is assessed by an ensemble of reward models, with each model providing a unique perspective on image restoration quality to guide the navigation process of the deterministic ODE trajectory. This creates a new paradigm for image restoration frameworks: dynamically trading computation for enhanced quality by exploiting the ODE structure.

We summarize the contributions of this work as follows:
\begin{itemize}
    \item \textbf{Flow-Based IR Framework:} We present the in-depth exploration of the FLUX flow model and its multimodal processing mechanism for image restoration, augmented by VLM-derived textual conditioning for pixel-aware and semantically-aware restoration.
    \item \textbf{ODE-adapted Test-Time Scaling:} We pioneer the adaptation of test-time scaling for ODE-based flow models image restoration framework, introducing a novel, training-free method to dynamically steer the deterministic sampling trajectory during the restoration process, establishing a new cost-performance trade-off.
    \item \textbf{State-of-the-Art Performance:} We validate through rigorous experimentation that our method surpasses existing diffusion-based and other approaches, providing a powerful new baseline for future research.
\end{itemize}

\section{Related Work}
\label{sec:formatting}

\textbf{Image Restoration with Pre-trained Generative Models.} Leveraging large-scale pre-trained generative models as priors to solve the ill-posed problem of image restoration has become a significant research trend. These methods aim to harness the powerful generative capabilities of such models to recover plausible and detailed high-quality images. A substantial body of research has been built upon Stable Diffusion (SD) models, such as StableSR \cite{wang2024exploiting}, DiffBIR \cite{lin2024diffbir}, and SeeSR \cite{wu2024seesr}. Building on this foundation, later studies have migrated to more powerful diffusion backbones \cite{podell2023sdxl, chen2023pixart}, including SUPIR \cite{yu2024scaling}, FaithDiff \cite{chen2025faithdiff}, and DreamClear \cite{ai2024dreamclear}. Recently, methods based on flow models \cite{martin2025pnp, xu2025fast, tsao2024boosting} have shown greater potential. Flow models learn a deterministic mapping from noise to data through Ordinary Differential Equations (ODEs), resulting in more coherent trajectories and often generating higher-quality samples with finer details. Building upon this frontier, we propose a novel image restoration framework based on the FLUX architecture. By integrating multimodal conditional guidance from both low-resolution images and semantic text, our framework delivers enriched contextual information throughout the restoration process.

\noindent \textbf{Test-Time Scaling for Generative Models.} The core idea of Test-Time Scaling (TTS) techniques is to dynamically mine and enhance the potential performance of pre-trained models during inference by investing additional computational resources, without modifying the model weights. This set of methods has achieved remarkable success in large language models  \cite{zhang2025survey}and is rapidly expanding into the visual generation domain. In the visual domain, the most straightforward TTS method is Best-of-N sampling \cite{stiennon2020learning, touvron2023llama, tang2024realfill}, which directly selects the sample with the highest reward signal from a large number of randomly generated candidates. Subsequently developed Particle Sampling methods  \cite{singh2025code, li2024derivative, kim2025test, li2025dynamic} retain high-performing sample branches and prune poorer ones during the denoising process, akin to beam search. However, their deterministic nature, due to the lack of stochasticity in the generative process, limits their applicability within more advanced diffusion and flow-based generative models. Furthermore, existing search methods heavily rely on the quality of the initial candidate particles and fail to actively explore new regions of the solution space. To address these limitations, we introduce a novel test-time extension paradigm, specifically designed for the ODE trajectories of flow models in image restoration. This training-free approach dynamically enhances output quality during inference, establishing a new computational-quality Pareto frontier for large models in low-level vision tasks.

\section{Method}

\begin{figure}[tb]
    \centering
    \includegraphics[width=\textwidth]{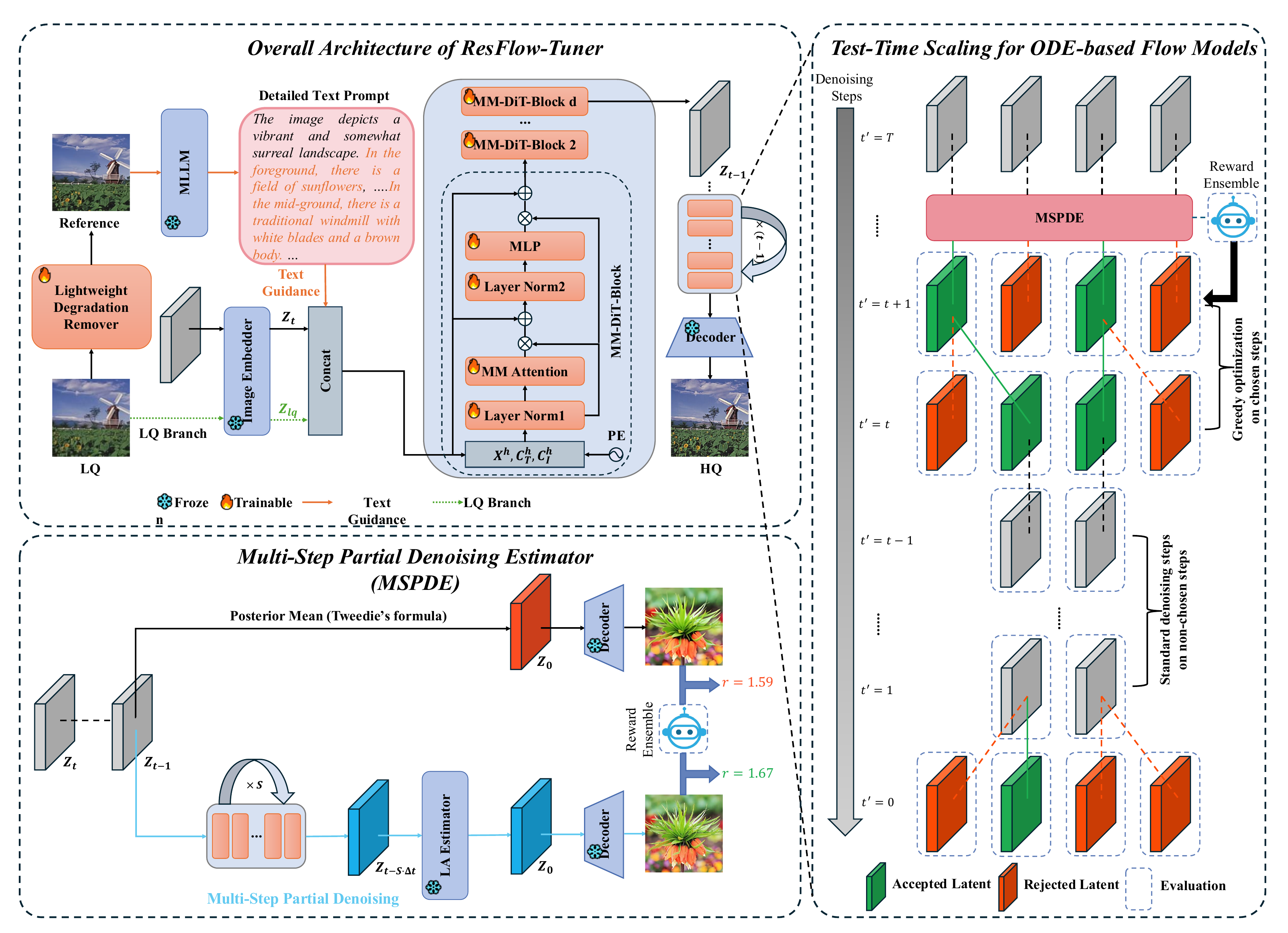}
    \caption{Architecture of the proposed \textbf{ResFlow-Tuner}. ResFlow-Tuner enhances training performance through the seamless integration of multi-modal guidance. During inference, it adopts a greedy optimization strategy for path selection, augmented by our Multi-Step Partial Denoising Estimator (MSPDE) for more accurate path evaluation.}
    \label{fig:2}
\end{figure}

\subsection{Flow-based Image Restoration Framework}
Our goal is to leverage the powerful generative prior of the FLUX model for image restoration, guided by both the LQ image and semantic text cues, As illustrated in Fig. \ref{fig:2}, our framework consists of two main components: (1) a condition preparation stage, comprising a lightweight degradation removal network and a multimodal large language model for generating textual descriptions; and (2) a conditional injection mechanism, specifically designed for our MM-DiT-based flow model to guide the image restoration process.

\subsubsection{Stage-I: Multi-Modal Condition Preparation}
\begin{itemize}
    \item \textbf{Reference Image Generation via SwinIR:} Directly processing the heavily degraded $I_{LR}$ with a Vision-Language Model (VLM) can lead to inaccurate semantic interpretations. To mitigate this, we first employ a pre-trained SwinIR model $\mathcal{G}_{SwinIR}$ to obtain a preliminary enhanced image $I_{Ref}$:
    \begin{equation}
    I_{Ref}=\mathcal{G}_{SwinIR}(I_{LR}).
    \end{equation}
    \item \textbf{Semantic Context Extraction via VLM:} The reference image $I_{Ref}$ is fed into a powerful VLM $\mathcal{F}_{VLM}$ (e.g., Qwen-2.5-VL) to generate a descriptive textual caption $T$ This caption encapsulates the high-level semantic content, serving as a weak textual condition to guide the generative process towards semantically coherent content synthesis.
    \begin{equation}
    T=\mathcal{F}_{VLM}(I_{Ref}).
    \end{equation}
\end{itemize}

\subsubsection{Stage-II: Unified Multi-Modal Fusion (UMMF) and ODE Solving}
% The FLUX model is built upon the Rectified Flow framework \cite{liu2022flow, liu2022rectified}, which learns a deterministic mapping between a simple noise distribution $p_1$ and the complex data distribution $p_0$ through a straight probability path. This is achieved by training a velocity field $v_t(z)$ to minimize the Flow Matching objective:

% % \begin{equation}
% % \mathcal{L}_{FM} = \mathbb{E}_{t,z_0 \sim p_0,z_1 \sim p_1,z_t \sim \mathcal{N}((1-t)z_0+tz_1,\sigma^2_1I)} \|v_\theta(z_t,t) - (z_1 - z_0)\|^2,
% % \end{equation}
% \begin{equation}
% \mathcal{L}_{FM} = \mathbb{E}_{t \sim U[0, 1], z_0 \sim p_0, z_1 \sim p_1} \|v_\theta(z_t, t) - (z_1 - z_0)\|^2
% \end{equation}
% where $z_t = (1 - t)z_0 + tz_1$ defines a linear trajectory. Once trained, sampling is performed by solving the ODE:

% \begin{equation}
% \frac{dz_t}{dt} = v_\theta(z_t,t),
% \end{equation}
% from $t = 1$ to $t = 0$. Our goal is to steer this flow so that it maps from the LR image distribution to the HR image distribution, conditioned on our multi-modal inputs.

\begin{itemize}
    \item \textbf{Unified Sequence Construction and Position-Aware Embedding:} Drawing from and adapting the multi-modal processing paradigm of advanced architectures like MM-DiT \cite{tan2025ominicontrol}, we employ a unified sequence concatenation approach for condition fusion, rather than using cross-attention. Specifically, we map the latent variable $z_{t}\in\mathbb{R}^{L_{z}\times d}$ representing the generative process into a token sequence. Concurrently, the low-quality image $I_{LQ}$ and the text caption $T$ are encoded into a visual token sequence $C_{img}=\mathcal{E}_{img}(I_{LQ})\in\mathbb{R}^{L_{img}\times d}$ and a text token sequence $C_{txt}=\mathcal{E}_{txt}(T)\in\mathbb{R}^{L_{txt}\times d}$, respectively.

    Subsequently, we concatenate these three components into a unified conditional generation sequence $\mathcal{S}$:
    
    \begin{equation}
    \mathcal{S}=[z_{t};C_{img};C_{txt}]\in\mathbb{R}^{(L_{z}+L_{img}+L_{txt})\times d}.
    \end{equation}
    To distinguish between different modalities and positional information within the sequence, we inject RoPE positional encoding $P$ and add a modality type encoding $M$ (e.g., $M_{latent},M_{image},M_{text}$) to each token. Thus, the final sequence input to the Transformer block is:

    \begin{equation}
    S^{(0)}=\mathcal{S}+P+M.
    \end{equation}
    \item \textbf{Forward Process of Concatenation-Based Multi-Modal DiT:} Condition injection no longer occurs externally to the attention layers but is naturally achieved through the self-attention mechanism within the unified sequence. Through this approach, the latent variable tokens $z_{t}$ can directly and adaptively retrieve and aggregate relevant information from the visual and textual condition tokens, achieving a deeper level of multi-modal fusion. The input sequence $S^{(0)}$ is processed by $N$ consecutive Transformer blocks. For the $\ell$-th block, the computation can be formally described as:

    \begin{equation}
    S^{(\ell)^{\prime}} =\text{LayerNorm}(S^{(\ell)}),
    \end{equation}
    \begin{equation}
    \mathcal{A}^{(\ell)} =S^{(\ell)^{\prime}}\text{SelfAttention}(S^{(\ell)^{\prime}})+S^{(\ell)},
    \end{equation}
    \begin{equation}
    S^{(\ell+1)} =S^{(\ell)^{\prime}}\text{MLP}(\text{LayerNorm}(\mathcal{A}^{(\ell)}))+S^{(\ell)}.
    \end{equation}
\end{itemize}

\subsection{Test-Time Scaling for ODE-based Flow Models}
While our two-stage image restoration framework produces high-quality results, we posit that the deterministic ODE trajectory of a flow model presents a unique and under-explored opportunity for inference-time optimization \cite{he2025scaling, kim2025inference}. By fundamentally reformulated for the ODE dynamics of flow models and the perceptual demands of image restoration, we introduce a Test-Time Scaling (TTS) paradigm.

\subsubsection{ODE-SDE Transformation Framework.}
To enable flow-based TTS, it is useful to revisit the theoretical connection between ODEs and SDEs. Flow models learn a velocity field vector, $u_{t}\in\mathbb{R}^{d}$, which defines an ODE trajectory. Sampling $x_0$ is then achieved by solving this ODE \cite{song2020score} in reverse time from $t=T$ to $t=0$:
\begin{equation}
\label{eq:ode}
x_{t-1}=x_{t}+u_{t}(x_{t})dt.
\end{equation}
This results in identical samples of $x_{t-1}$ being derived from any given $x_t$, implying a deterministic mapping. Consequently, the applicability of test-time extension methods to flow models is limited, as the sampling process lacks inherent stochasticity beyond the initial noise. To overcome this constraint, we transform the deterministic Flow-ODE into an equivalent Stochastic Differential Equation (SDE) by introducing a stochastic diffusion term. Following prior work \cite{albergo2023stochastic, kim2025inference, ma2024sit, patel2024exploring, singh2024stochastic}, the ODE sampling process in Eq. \ref{eq:ode} is reformulated as follows:
\begin{equation}
\label{eq:sde}
dx_{t}=\left(u_{t}(x_{t})-\frac{\sigma^{2}_{t}}{2}\nabla \log p_{t}(x_{t})\right)dt+\sigma_{t}dw.
\end{equation}
In this formulation, the score $\log p_{t}(x_{t})$ can be computed from the velocity field $u_t$ \cite{singh2024stochastic}, while the Wiener process $dw$ injects stochasticity at each sampling step. This SDE framework thereby introduces randomness into the trajectory. Building upon this, our method operates directly on the deterministic ODE trajectory of the FLUX model and performs its search by guiding perturbations with this injected randomness, effectively leveraging the inherently straighter paths characteristic of ODEs.

An ODE, by construction, defines a deterministic mapping: given an initial noise and conditioning input, its output is uniquely determined. Arbitrarily perturbing an intermediate state would disrupt this learned deterministic dynamics, potentially causing the resulting state to deviate from the model's learned data manifold and leading to quality degradation or semantic errors. Critically, the ODE-to-SDE theory provides the theoretical foundation for our approach. In essence, this theory establishes that for a deterministic ODE defined in the form of Eq. \ref{eq:ode}, one can construct a family of SDEs as in Eq. \ref{eq:sde}. By designing their drift coefficients (set to $\sigma_t$ in our paper) and introducing stochastic noise $dw$, the resulting stochastic processes share the same marginal distributions as the corresponding ODE model but exhibit diverse sampling trajectories. The “mutation” operation we introduce next is precisely designed to explore within this legitimate SDE family, rather than arbitrarily disrupting the model. Our objective is not to compromise the distribution, but rather to explore—while preserving the marginal distributions—for samples that yield superior outcomes according to our reward function (\textit{i.e.}, image quality). The SDE theory thereby delineates the search space for our "legitimate exploration." However, since our algorithmic goal is not general stochastic sampling but rather injecting controlled stochasticity into the pre-trained, highly deterministic FLUX ODE trajectory to explore better image restoration paths, we specifically apply and discretize Eq. \ref{eq:sde} as shown in Eq. \ref{eq:perturbation} and Eq. \ref{eq:MSPDE} below. The principles underlying this discretization and the logical connections behind it will be detailed in the corresponding section of a later chapter.

\subsubsection{Formalizing TTS as Trajectory Optimization in Latent Space.}
The core intuition is to treat the initial state $z_1$ of the ODE as a searchable variable within a neighborhood, and the ODE solver as a dynamical system whose trajectory can be steered. Unlike diffusion models with stochastic differential equations (SDEs), the ODE of a flow model like FLUX defines a deterministic mapping $\Phi_{1 \to 0}(z_1; c)$ from initial noise $z_1$ to the final latent $z_0$, given condition $c$. Our goal is to find an initial state $z_1^*$ within a constrained set $\mathcal{Z}$ that maximizes the expected quality of the resulting image:
\begin{equation}
    z_1^* = \arg \max_{z_i \in \mathcal{Z}} \mathbb{E} \left[ \mathcal{R}(\mathcal{D}(\Phi_{1 \to 0}(z_1; c))) \right],
\end{equation}
where $\mathcal{R}$ is a reward function quantifying image quality, and $\mathcal{D}$ is the decoder. Directly solving this optimization over the high-dimensional $z_1$ is intractable. Instead, we perform an iterative, greedy optimization over the trajectory at $K$ strategically chosen time steps $\{t_k\}_{k=1}^K$, where $1 = t_1 > t_2 > \cdots > t_K > 0$.

\begin{figure}[tb]
    \centering
    \captionsetup[subfigure]{font=tiny}
    
    % 重新计算子图宽度：7个子图，考虑间距
    \begin{subfigure}[b]{0.135\textwidth}
        \centering
        \includegraphics[width=\textwidth]{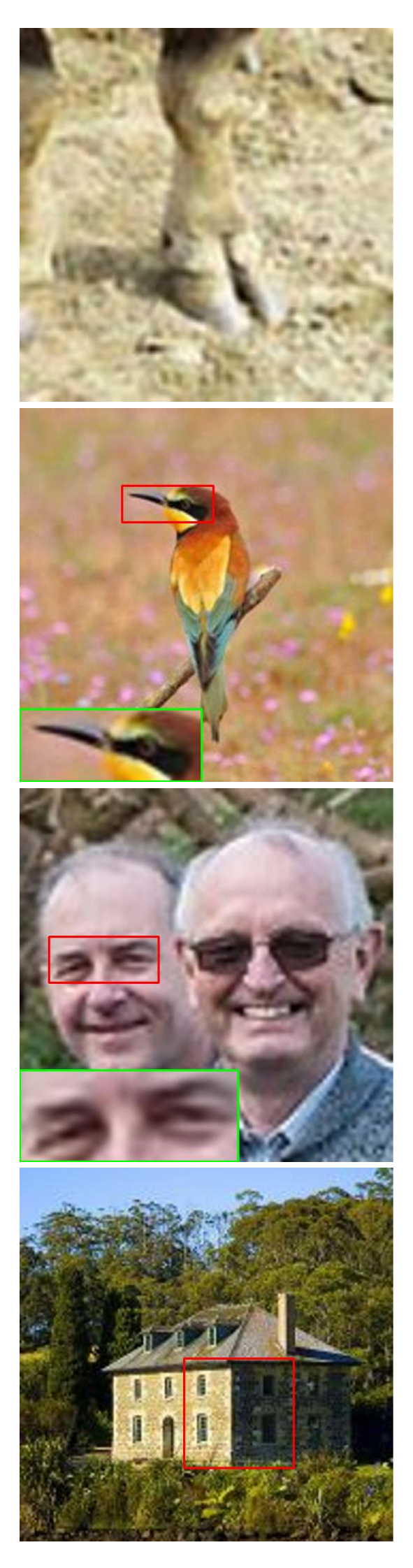}
        \caption{LQ Input}
    \end{subfigure}
    \hspace{-0.8em} % 调整负间距
    \begin{subfigure}[b]{0.135\textwidth}
        \centering
        \includegraphics[width=\textwidth]{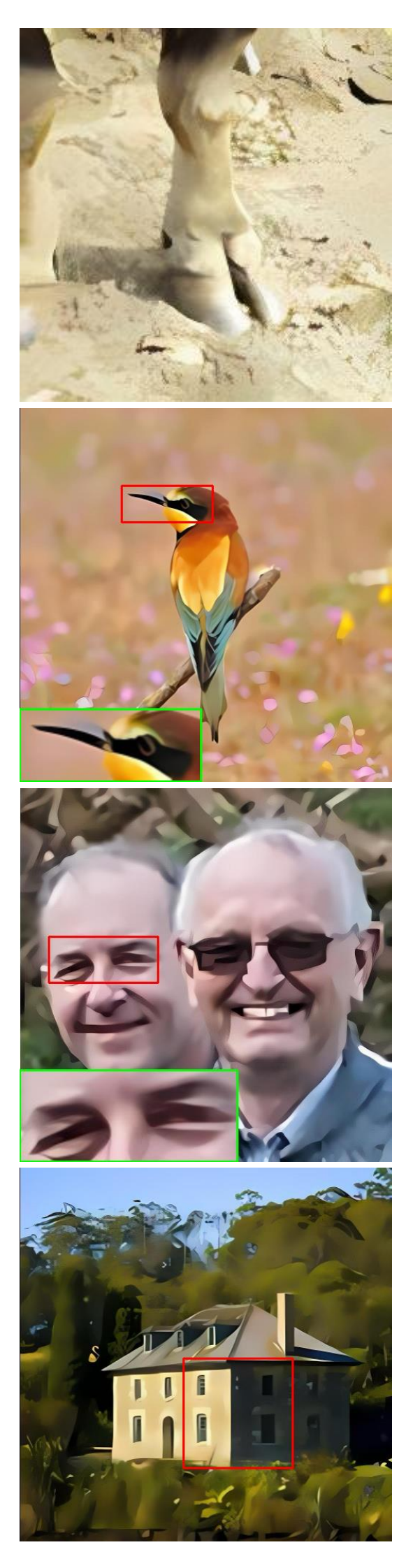}
        \caption{Real-ESR}
    \end{subfigure}
    \hspace{-0.8em}
    \begin{subfigure}[b]{0.135\textwidth}
        \centering
        \includegraphics[width=\textwidth]{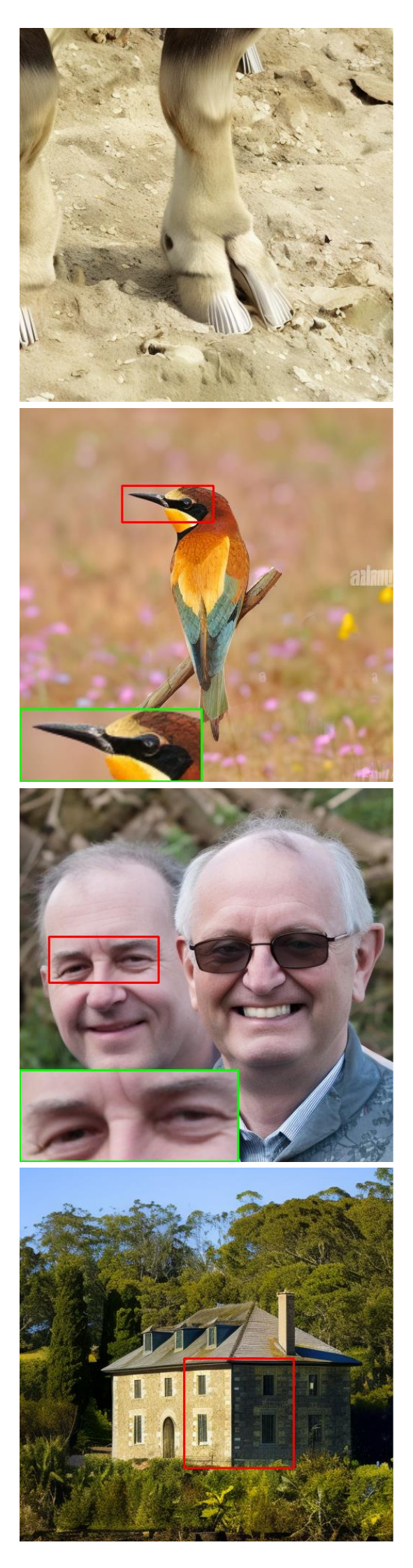}
        \caption{DiffBIR}
    \end{subfigure}
    \hspace{-0.8em}
    \begin{subfigure}[b]{0.135\textwidth}
        \centering
        \includegraphics[width=\textwidth]{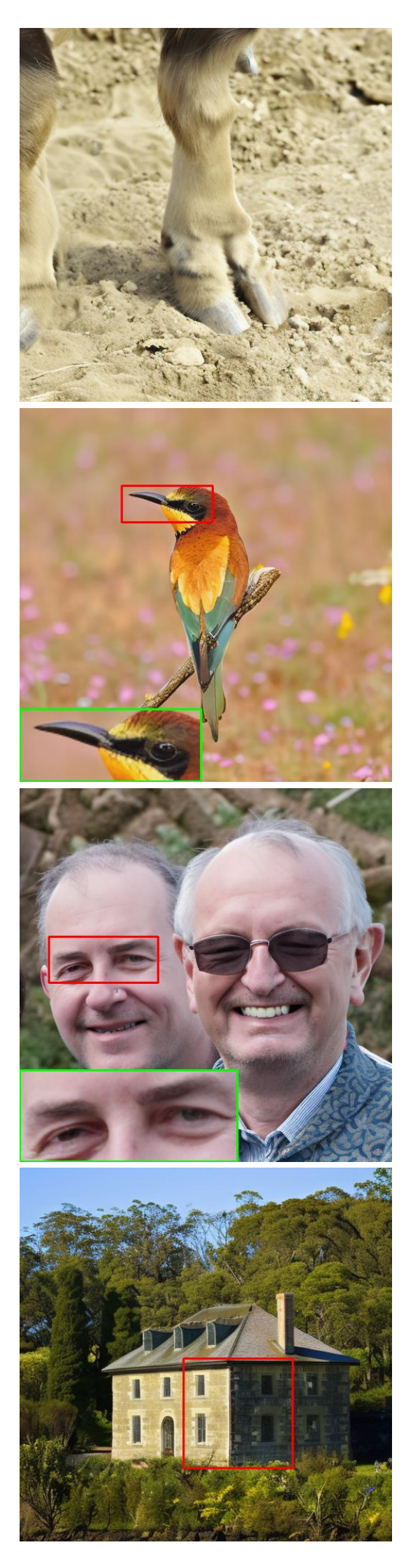}
        \caption{SeeSR}
    \end{subfigure}
    \hspace{-0.8em}
    \begin{subfigure}[b]{0.135\textwidth}
        \centering
        \includegraphics[width=\textwidth]{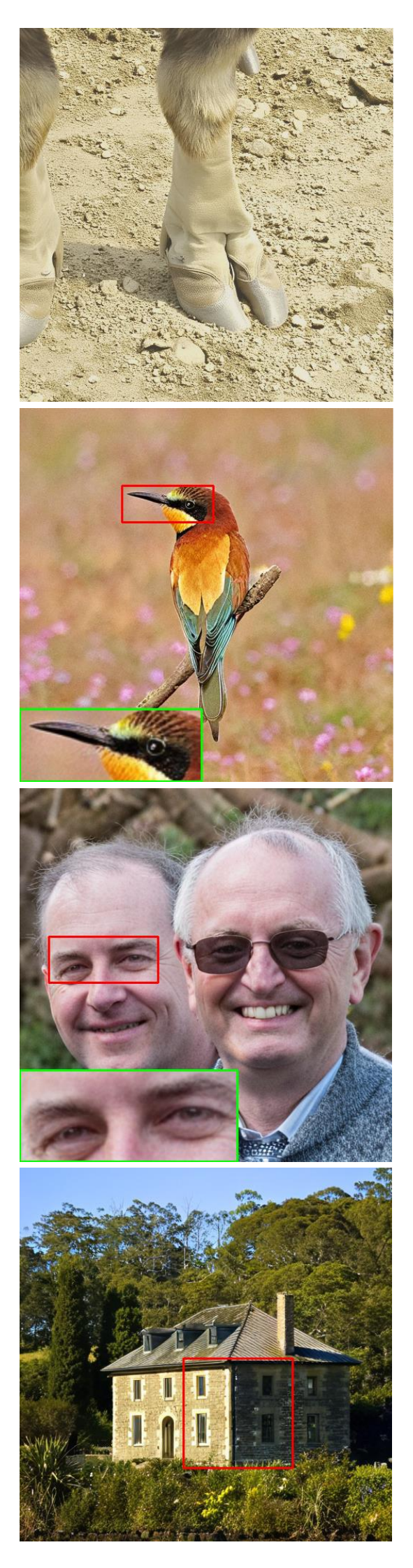}
        \caption{SUPIR}
    \end{subfigure}
    \hspace{-0.8em}
    \begin{subfigure}[b]{0.135\textwidth}
        \centering
        \includegraphics[width=\textwidth]{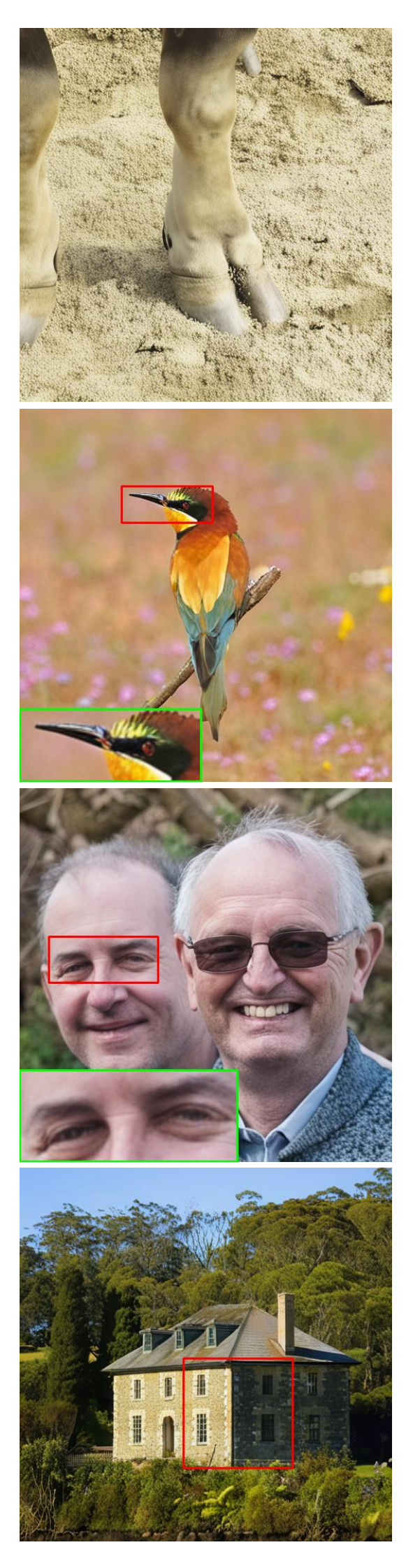}
        \caption{FaithDiff}
    \end{subfigure}
    \hspace{-0.8em}
    \begin{subfigure}[b]{0.135\textwidth}
        \centering
        \includegraphics[width=\textwidth]{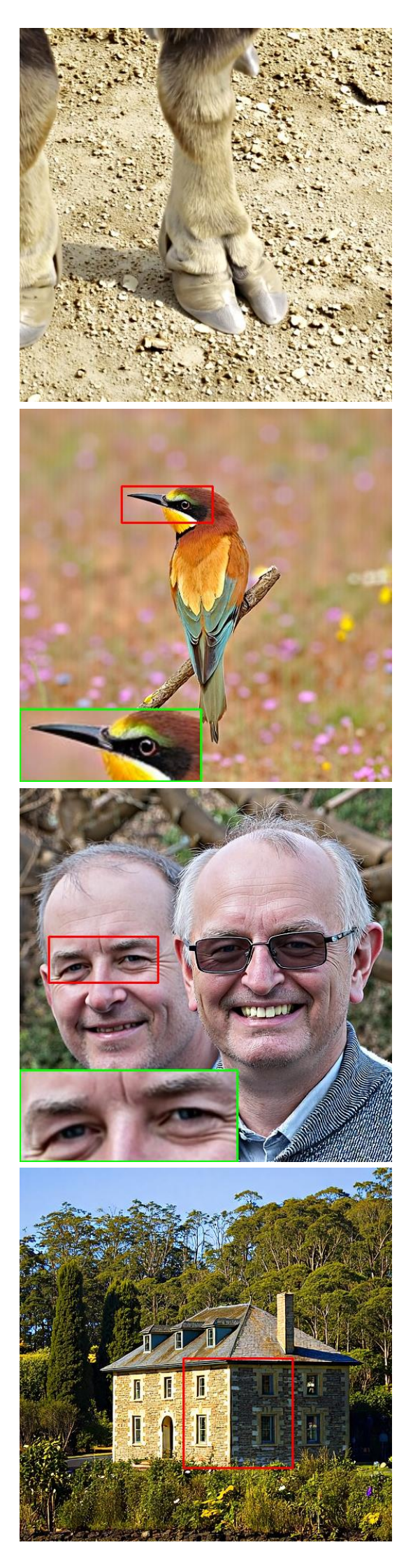}
        \caption{Ours}
    \end{subfigure}
    
    \caption{Qualitative comparisons on both synthetic (the first row) and real-world (the last three rows) benchmarks. Please zoom in for a better view.}
    \label{fig:3}
\end{figure}

\subsubsection{ODE-aware Search with Verifier Ensemble.}
At each intervention step $t_k$, let the current state be $z_{t_k}$. The algorithm generates $N$ candidate perturbations of the current state, evaluates their potential future quality via a partial ODE rollout and a learned reward model, and selects the most promising path. This "perturb-rollout-evaluate-select" cycle is detailed below:
\begin{enumerate}
    \item \textbf{Perturbation}: We generate $N$ candidate perturbations $\{\bar{z}_{t_{k}}^{(i)}\}_{i=1}^{N}$ by injecting structured noise:
    \begin{equation}
    \label{eq:perturbation}
        \bar{z}_{t_{k}}^{(i)}=z_{t_{k}}+\epsilon^{(i)},\quad\epsilon^{(i)} \sim\mathcal{N}(0,\sigma_{k}^{2}I).
    \end{equation}
    This is equivalent to $\bar{z}_{t_k}^{(i)} = z_{t_k} + \sigma_k^{(i)} \epsilon_k^{(i)}, \quad \epsilon_k^{(i)} \sim \mathcal{N}(0, I)$, an alternative mutation operator we propose, inspired by the reverse-time SDE, to synthesize meaningful variations while preserving the intrinsic structure of the latent state $z_{t_{k}}$.
    
    Here, $\sigma_k$ is the diffusion coefficient defined in the reverse-time SDE (Eq. \ref{eq:sde}), which controls the level of injected stochasticity and can be scheduled to decrease with $k$, encouraging coarse exploration early and fine-tuning later. This mutation operation effectively generates novel latent states, enabling exploration of an expanded state space while preserving the inherent distribution established during the denoising process. In the next generation of our TTS search, we sample $z_0^{(i)} \sim p_0(z_0^{(i)} \mid \bar{z}_{t_k}^{(i)})$ based on the new offspring $\bar{z}_{t_k}^{(i)}$ and repeat the above evolutionary search process, including evaluation, selection, and mutation. Therefore, $\epsilon^{(i)}$ here in Eq. \ref{eq:perturbation} is essentially equivalent to the Wiener process $\sigma_t dw$ in Eq. \ref{eq:sde}.
    \item \textbf{Accurate $z_0$ Estimation via Multi-Step Partial Denoising (MSPDE)}: We obtain partially denoised latent variables by performing a fixed number of ODE solving steps, then decode and evaluate them. Given a candidate perturbation $\tilde{z}_{t_{k}}^{(i)}$, we perform $S$ steps of ODE solving to obtain the partially denoised latent variable:
    \begin{equation}
    \label{eq:MSPDE}
        \tilde{z}_{t_{k}-S\cdot\Delta t}^{(i)}=\text{ODESolver}(\tilde{z}_{t_{k}}^{(i)}, t_{k},t_{k}-S\cdot\Delta t,\mathbf{c})
    \end{equation}
    where $\Delta t$ is the single-step time interval and $S$ is the fixed number of denoising steps. We then use a a lookahead estimator \cite{oshima2025inference}to obtain a high-fidelity preview of the final image $\tilde{z}_0^{(i)}$, rather than decoding a noisy intermediate latent state. This method balances computational efficiency with high assessment accuracy, mitigating the error inherent in one-step estimators.

    Eq.~\ref{eq:MSPDE} sets the drift term in the SDE (\textit{i.e.}, $\frac{\sigma_t^2}{2} \nabla \log p_t(x_t) dt$ in Eq.~\ref{eq:sde}) to zero—a key design choice in our framework. In the continuous SDE formulation shown in Eq.~\ref{eq:sde}, $\nabla \log p_t(x_t)$ serves as a score function that guides the sampling process toward the data distribution $p_t(x_t)$, representing a general, task-agnostic gradient. However, image restoration is inherently a conditional generation problem that prioritizes local optimization within the output space over exact simulation of the full data distribution. Unlike unconditional text-to-image tasks that pursue broad distribution coverage and diversity (e.g., as measured by FID), our objective is to recover a restored image that is optimal in terms of pixel fidelity, perceptual quality, and semantic consistency. This constitutes an optimization problem with a well-defined objective function. Accordingly, we employ a Verifier Ensemble to steer the search direction toward human visual preferences and semantic realism, rather than merely toward the original data distribution. This conceptual shift—from a \textit{data distribution score} to a \textit{task-specific reward signal}—represents a core innovation in adapting SDE-based generative principles to discriminative, optimization-driven tasks such as image restoration. Our results demonstrate that even without strictly adhering to the continuous SDE formalism, the introduction of discrete stochasticity can effectively correct biases (e.g., variance underestimation) that arise from discretization errors in purely deterministic ODE solvers.
    
    Our strategy can thus be interpreted as follows: at sparse yet critical time steps, we harness the stochasticity inherent in SDE theory to "jump" to the starting point of a new latent trajectory. From there, we perform a fast, low-cost rollout evaluation along the corresponding deterministic ODE path originating from that point. Notably, this ODE path itself corresponds to a specific instance of the original SDE family, obtained by setting its drift term to zero.
    \item \textbf{Verifier Ensemble for Image Restoration}: Our reward function $\mathcal{R}$ is designed tailored for perceptual image restoration. Instead of relying on a single metric, we form a Verifier Ensemble $\mathcal{R}_{ensemble}$ that synthesizes judgments from multiple perspectives (\textit{i.e.}, Aesthetic Score Predictor \cite{schuhmann2022laion}, CLIPScore \cite{hessel2021clipscore} and ImageReward \cite{xu2023imagereward}). Since these validators operate at different scales, we adopt a rank-based ensemble strategy. Specifically, for the $N$ candidates at a given step, we compute the ordinal rank from each validator—denoted as $r_{aes}^{(i)}$, $r_{clip}^{(i)}$ and $r_{ir}^{(i)}$ for candidate $i$.
    \begin{equation}
        \mathcal{R}_{ensemble}(\hat{I}^{(i)}) = -\frac{1}{3} \left( r_{aes}^{(i)} + r_{clip}^{(i)} + r_{ir}^{(i)} \right)
    \end{equation}
    This approach favors candidates that demonstrate consistent performance across all quality dimensions, making it inherently robust to scale discrepancies.
    \item \textbf{Selection and Commitment}: Instead of a simple "winner-takes-all" strategy, we adopt a progressive filtering approach inspired by EvoSearch \cite{he2025scaling}. The $N$ candidates are ranked by their ensemble reward. We then retain the top-$M$ performers (where $M<N$), and these form the parent population for the next iteration, which is chosen as the starting point for the immediate next ODE segment. This strategy maintains diversity in the search space for longer, increasing the probability of finding a globally superior path.
\end{enumerate}

\vspace{-0.5cm}
\begin{algorithm}
\renewcommand{\thealgorithm}{}
\caption{ODE-adapted Test-Time Scaling}
\label{alg:tts-flow-concise}
\begin{algorithmic}[1]
\Require Pre-trained flow IR model $\Phi$, condition $c$, initial noise $z_1$
\Ensure Optimized restored image $\hat{x}_0$
\State Initialize population $\mathcal{P} \gets \{z_1\}$
\For{$k = 1$ \textbf{to} $K$} \Comment{$K$ intervention steps}
    \State $\mathcal{C} \gets \text{Perturb}(\mathcal{P}, \sigma_k)$ \Comment{Generate $N$ candidates with noise $\sigma_k$}
    
    \For{each $z^{(i)} \in \mathcal{C}$}
        \State $\hat{z}_0^{(i)} \gets \text{MSPDE}(z^{(i)}, t_k, S)$ \Comment{Multi-step partial denoising}
        \State $r^{(i)} \gets \mathcal{R}_{\text{ensemble}}(\mathcal{D}(\hat{z}_0^{(i)}))$ \Comment{Rank-based reward ensemble}
    \EndFor
    
    \State $\mathcal{P} \gets \text{TopM}(\mathcal{C}, r, M)$ \Comment{Select top-$M$ candidates}
    \State $\mathcal{P} \gets \text{AdvanceODE}(\mathcal{P}, t_k, t_{k+1}, c)$ \Comment{Move to next time step}
\EndFor

\State $\hat{x}_0 \gets \mathcal{D}(\text{ODESolver}(\mathcal{P}[0], t_K, 0, c))$ \Comment{Final denoising}
\State \Return $\hat{x}_0$
\end{algorithmic}
\end{algorithm}
\vspace{-1cm}

\section{Experiments}

\subsection{Experimental Setup}
\label{sec:Experimental Setup}

\textbf{Training Dataset:} Our model was trained on a combination of the DIV2K \cite{lim2017enhanced}, Flickr2K \cite{agustsson2017ntire}, LSDIR \cite{li2023lsdir}, DIV8K \cite{gu2019div8k}. Following common practices in Real-ESRGAN \cite{wang2021real}, we generated LQ counterparts by applying a bicubic downsampling (with a scaling factor of ×4) to the HQ images, using the same degradation settings as StableSR \cite{wang2024exploiting} to ensure a fair comparison.

\noindent \textbf{Testing Datasets:} To comprehensively evaluate the generalization ability and performance of our method, we conducted tests both on synthetic test datasets and real-world test datasets. For synthetic benchmarks, we employ different levels of degradations (D-level) to synthesize degraded validation sets of DIV2K-Val, similar to DASR \cite{liang2022efficient}, and degrade LSDIR-Val using the same settings as training. For real-world benchmarks, we conduct experiments on commonly used RealPhoto60\cite{yu2024scaling}, RealSR  \cite{cai2019toward} and DRealSR \cite{wei2020component} datasets.

\noindent \textbf{Implementation Details:} We build our method upon FLUX.1 \cite{flux2024} and Qwen2.5-VL \cite{Qwen2.5-VL}. By default, we use FLUX.1-dev to generate images based on given low-quality inputs and Qwen2.5-VL-7B to generate captions based on reference images. Our method utilizes LoRA \cite{hu2022lora} for fine-tuning the base model with a default rank of 16. The SwinIR model in DiffBIR \cite{lin2024diffbir} is used as the lightweight degradation remover. Our model is trained on $512\times 512$ resolution images, with a batch size of 16 and gradient accumulation over 4 batch (effective batch size of 64). We employ the Prodigy optimizer \cite{mishchenko2023prodigy} with safeguard warmup and bias correction enabled, setting the weight decay to 0.01. Models are trained for 30,000 iterations. For inference of our model, we employ the sde-dpmsolver++ sampler in FlowDPMSolverMultistepScheduler \cite{von-platen-etal-2022-diffusers} in SDE process with 50 denoising steps and set guidance scale as 5.5. We set the mutation rate $\beta$ = 0.3, with $\sigma_t$ following the default sde-dpmsolver configurations.

\begin{table}[tb]
\caption{Quantitative comparison with state-of-the-art methods on synthetic benchmark. ‘I’, ‘II’, and ‘III’ denote the datasets with mild, medium, and severe degradations, respectively. The best and second performances are marked in \textcolor{red}{\textbf{red}} and \textcolor{blue}{\textbf{blue}}, respectively.}
\label{tab:synthetic_comparison}
\centering
\resizebox{\textwidth}{!}{
\renewcommand{\arraystretch}{0.7}
\footnotesize
\setlength{\tabcolsep}{2pt}
\begin{tabular}{@{}cc*{14}{c}@{}}
\toprule
\multirow{2}{*}{\textbf{Dataset}} & \multirow{2}{*}{\textbf{Metric}} & \textbf{Real-ESRGAN} & \textbf{BSRGAN} & \textbf{StableSR} & \textbf{DiffBIR} & \textbf{PASD} & \textbf{SeeSR} & \textbf{DreamClear} & \textbf{OSEDiff} & \textbf{SUPIR} & \textbf{FaithDiff} & \textbf{PnP-Flow} & \textbf{FlowSR} & \textbf{LP} & \textbf{Ours}\\
& & \cite{wang2021real} & \cite{zhang2021designing} & \cite{wang2024exploiting} & \cite{lin2024diffbir} & \cite{yang2024pixel} & \cite{wu2024seesr} & \cite{ai2024dreamclear} & \cite{wu2024one} & \cite{yu2024scaling} & \cite{chen2025faithdiff} & \cite{martin2025pnp} & \cite{xu2025fast} & \cite{tsao2024boosting} &  \\
\midrule
\multirow{7}{*}{\rotatebox[origin=c]{90}{\scriptsize\textbf{DIV2K-Val-I}}}
& PSNR$\,\textcolor{red}{\uparrow}$ & \textcolor{blue}{\textbf{26.64}} & \textcolor{red}{\textbf{27.63}} & 24.71 & 24.60 & 25.31 & 25.08 & 23.76 & 25.13 & 25.09 & 24.29 & 23.82 & 25.02 & 22.98 & 23.11 \\

& SSIM$\,\textcolor{red}{\uparrow}$ & \textcolor{blue}{\textbf{0.7737}} & \textcolor{red}{\textbf{0.7897}} & 0.7131 & 0.6595 & 0.6995 & 0.6967 & 0.6574 & 0.7098 & 0.7010 & 0.6668 & 0.6632 & 0.7219 & 0.6932 & 0.6945 \\

& LPIPS$\,\textcolor{red}{\downarrow}$ & \textcolor{red}{\textbf{0.1964}} & \textcolor{blue}{\textbf{0.2038}} & 0.2393 & 0.2496 & 0.2370 & 0.2263 & 0.2259 & 0.2382 & 0.2139 & 0.2187 & 0.2214 & 0.2143 & 0.2198 & 0.2098 \\

& PaQ-2-PiQ$\,\textcolor{red}{\uparrow}$ & 73.85 & 74.17 & 74.19 & 72.81 & 73.12 & 73.92 & 73.86 & 74.07 & 74.24 & \textcolor{blue}{\textbf{74.76}} & 73.21 & 74.14 & 73.82 & \textcolor{red}{\textbf{78.82}} \\

& MANIQA$\,\textcolor{red}{\uparrow}$ & 0.5611 & 0.5507 & 0.5743 & 0.5898 & 0.5524 & 0.5749 & 0.5750 & 0.5767 & \textcolor{blue}{\textbf{0.6049}} & 0.5945 & 0.5511 & 0.5828 & 0.5842 & \textcolor{red}{\textbf{0.7261}} \\

& MUSIQ$\,\textcolor{red}{\uparrow}$ & 62.38 & 61.81 & 65.55 & 66.23 & 64.57 & 66.48 & 66.15 & 66.39 & 65.49 & \textcolor{blue}{\textbf{66.53}} & 65.44 & 66.24 & 64.85 & \textcolor{red}{\textbf{75.54}} \\

& CLIPIQA+$\,\textcolor{red}{\uparrow}$ & 0.4649 & 0.4588 & 0.5156 & 0.5407 & 0.4764 & 0.5336 & \textcolor{blue}{\textbf{0.5478}} & 0.5271 & 0.5202 & 0.5432 & 0.5294 & 0.5328 & 0.5422 & \textcolor{red}{\textbf{0.7925}} \\
\midrule
\multirow{7}{*}{\rotatebox[origin=c]{90}{\scriptsize\textbf{DIV2K-Val-II}}}
& PSNR$\,\textcolor{red}{\uparrow}$ & \textcolor{blue}{\textbf{25.49}} & \textcolor{red}{\textbf{26.42}} & 24.26 & 24.42 & 24.89 & 24.65 & 23.39 & 25.12 & 24.42 & 23.80 & 23.45 & 24.79 & 22.54 & 22.79 \\

& SSIM$\,\textcolor{red}{\uparrow}$ & \textcolor{blue}{\textbf{0.7274}} & \textcolor{red}{\textbf{0.7402}} & 0.6771 & 0.6441 & 0.6764 & 0.6734 & 0.6330 & 0.6983 & 0.6703 & 0.6413 & 0.6421 & 0.6983 & 0.6708 & 0.6778 \\

& LPIPS$\,\textcolor{red}{\downarrow}$ & 0.2309 & 0.2465 & 0.2590 & 0.2708 & 0.2502 & 0.2428 & 0.2518 & 0.2499 & 0.2432 & 0.2407 & 0.2398 & \textcolor{blue}{\textbf{0.2303}} & 0.2346 & \textcolor{red}{\textbf{0.2255}} \\

& PaQ-2-PiQ$\,\textcolor{red}{\uparrow}$ & 73.39 & 73.26 & 73.81 & 72.07 & 72.91 & 73.52 & 73.28 & 73.67 & 74.26 & \textcolor{blue}{\textbf{74.77}} & 73.05 & 73.82 & 73.54 & \textcolor{red}{\textbf{78.94}} \\

& MANIQA$\,\textcolor{red}{\uparrow}$ & 0.5499 & 0.5261 & 0.5825 & 0.5827 & 0.5466 & 0.5691 & 0.5667 & 0.5675 & \textcolor{blue}{\textbf{0.6012}} & 0.5919 & 0.5342 & 0.5682 & 0.5624 & \textcolor{red}{\textbf{0.7038}} \\

& MUSIQ$\,\textcolor{red}{\uparrow}$ & 61.84 & 60.00 & 64.76 & 64.83 & 64.45 & 66.09 & 64.96 & 65.04 & 65.58 & \textcolor{blue}{\textbf{66.42}} & 64.31 & 65.08 & 63.98 & \textcolor{red}{\textbf{74.68}} \\

& CLIPIQA+$\,\textcolor{red}{\uparrow}$ & 0.4719 & 0.4463 & 0.5057 & 0.5246 & 0.4718 & 0.5226 & 0.5295 & 0.5209 & 0.5202 & \textcolor{blue}{\textbf{0.5460}} & 0.5188 & 0.5204 & 0.5342 & \textcolor{red}{\textbf{0.7811}} \\
\midrule
\multirow{7}{*}{\rotatebox[origin=c]{90}{\scriptsize\textbf{DIV2K-Val-III}}}
& PSNR$\,\textcolor{red}{\uparrow}$ & 22.81 & \textcolor{blue}{\textbf{23.45}} & 23.34 & 23.42 & 22.58 & 22.58 & 21.82 & \textcolor{red}{\textbf{24.10}} & 21.90 & 21.77 & 20.22 & 21.53 & 19.48 & 19.74 \\

& SSIM$\,\textcolor{red}{\uparrow}$ & \textcolor{blue}{\textbf{0.6288}} & 0.6281 & 0.6277 & 0.5992 & 0.5985 & 0.5944 & 0.5510 & \textcolor{red}{\textbf{0.6432}} & 0.5611 & 0.5662 & 0.5034 & 0.5536 & 0.5286 & 0.5263 \\

& LPIPS$\,\textcolor{red}{\downarrow}$ & 0.3535 & 0.3462 & 0.3559 & 0.3676 & 0.3646 & 0.3278 & 0.3336 & \textcolor{red}{\textbf{0.2978}} & 0.3172 & 0.3080 & 0.3222 & 0.3182 & 0.3228 & \textcolor{blue}{\textbf{0.3047}} \\

& PaQ-2-PiQ$\,\textcolor{red}{\uparrow}$ & 71.93 & 72.09 & 70.22 & 74.20 & 72.68 & 73.46 & 72.49 & 73.49 & 74.07 & \textcolor{blue}{\textbf{75.05}} & 73.12 & 73.48 & 73.31 & \textcolor{red}{\textbf{78.62}} \\

& MANIQA$\,\textcolor{red}{\uparrow}$ & 0.5057 & 0.4917 & 0.4896 & 0.5776 & 0.4999 & 0.5462 & 0.5515 & 0.5405 & \textcolor{blue}{\textbf{0.5882}} & 0.5686 & 0.4976 & 0.5206 & 0.5223 & \textcolor{red}{\textbf{0.6605}} \\

& MUSIQ$\,\textcolor{red}{\uparrow}$ & 60.11 & 62.41 & 57.89 & 58.86 & 63.08 & 65.82 & 62.59 & 65.28 & 65.46 & \textcolor{blue}{\textbf{66.28}} & 64.12 & 64.88 & 63.54 & \textcolor{red}{\textbf{74.67}} \\

& CLIPIQA+$\,\textcolor{red}{\uparrow}$ & 0.4637 & 0.4838 & 0.4124 & 0.5154 & 0.4815 & 0.5106 & 0.4914 & 0.5132 & 0.5134 & 0.5275 & 0.5072 & 0.5126 & 0.5225 & \textcolor{red}{\textbf{0.7704}} \\
\midrule
\multirow{8}{*}{\rotatebox[origin=c]{90}{\scriptsize\textbf{LSDIR-Val}}}
& PSNR$\,\textcolor{red}{\uparrow}$ & 18.13 & 18.27 & 18.11 & \textcolor{blue}{\textbf{18.42}} & 18.21 & 18.03 & 17.01 & \textcolor{red}{\textbf{19.02}} & 16.95 & 16.87 & 15.98 & 16.82 & 14.82 & 15.02 \\

& SSIM$\,\textcolor{red}{\uparrow}$ & \textcolor{red}{\textbf{0.4867}} & 0.4673 & 0.4508 & 0.4618 & 0.4592 & 0.4564 & 0.4236 & \textcolor{blue}{\textbf{0.4796}} & 0.4080 & 0.4259 & 0.3802 & 0.4044 & 0.3907 & 0.3845 \\

& LPIPS$\,\textcolor{red}{\downarrow}$ & 0.3986 & 0.4378 & 0.4152 & 0.4049 & 0.3883 & 0.3759 & 0.3836 & \textcolor{blue}{\textbf{0.3742}} & 0.4119 & 0.3802 & 0.3954 & 0.3822 & 0.3810 & \textcolor{red}{\textbf{0.3721}} \\

& DISTS$\,\textcolor{red}{\downarrow}$ & 0.2278 & 0.2539 & 0.2159 & 0.2439 & 0.1734 & 0.1966 & 0.1656 & 0.1619 & 0.1838 & \textcolor{blue}{\textbf{0.1557}} & 0.1675 & 0.1582 & 0.1653 & \textcolor{red}{\textbf{0.1406}} \\

& FID$\,\textcolor{red}{\downarrow}$ & 46.46 & 53.25 & 31.26 & 35.91 & 22.65 & 25.91 & 22.06 & 22.32 & 30.03 & \textcolor{blue}{\textbf{20.69}} & 22.93 & 22.18 & 22.76 & \textcolor{red}{\textbf{19.82}} \\

& MANIQA$\,\textcolor{red}{\uparrow}$ & 0.4381 & 0.3829 & 0.3098 & 0.4551 & 0.4043 & 0.5700 & 0.4811 & 0.4630 & 0.4683 & 0.5956 & 0.4878 & \textcolor{blue}{\textbf{0.6023}} & 0.5912 & \textcolor{red}{\textbf{0.6658}} \\

& MUSIQ$\,\textcolor{red}{\uparrow}$ & 68.25 & 65.98 & 59.37 & 65.94 & 71.02 & \textcolor{blue}{\textbf{73.00}} & 70.40 & 70.74 & 70.98 & 72.64 & 71.15 & 71.68 & 70.82 & \textcolor{red}{\textbf{74.21}} \\

& CLIPIQA$\,\textcolor{red}{\uparrow}$ & 0.6218 & 0.5648 & 0.5190 & 0.6592 & 0.5729 & \textcolor{blue}{\textbf{0.7261}} & 0.6914 & 0.6679 & 0.6174 & 0.7224 & 0.7152 & 0.7208 & 0.7097 & \textcolor{red}{\textbf{0.7389}} \\
\bottomrule
\end{tabular}
}
\end{table}
\vspace{-0.5cm}

\begin{table}[tb]
\caption{Quantitative comparison with state-of-the-art methods on real-world benchmark. The best and second performances are marked in \textcolor{red}{\textbf{red}} and \textcolor{blue}{\textbf{blue}}, respectively.}
\label{tab:real_comparison}
\centering
\resizebox{\textwidth}{!}{
\renewcommand{\arraystretch}{0.7}
\footnotesize
\setlength{\tabcolsep}{3pt}
\begin{tabular}{@{}cc*{14}{c}@{}}
\toprule
\multirow{2}{*}{\textbf{Dataset}} & \multirow{2}{*}{\textbf{Metric}} & \textbf{Real-ESRGAN} & \textbf{BSRGAN} & \textbf{StableSR} & \textbf{DiffBIR} & \textbf{PASD} & \textbf{SeeSR} & \textbf{DreamClear} & \textbf{OSEDiff} & \textbf{SUPIR} & \textbf{FaithDiff} & \textbf{PnP-Flow} & \textbf{FlowSR} & \textbf{LP} & \textbf{Ours}\\
& & \cite{wang2021real} & \cite{zhang2021designing} & \cite{wang2024exploiting} & \cite{lin2024diffbir} & \cite{yang2024pixel} & \cite{wu2024seesr} & \cite{ai2024dreamclear} & \cite{wu2024one} & \cite{yu2024scaling} & \cite{chen2025faithdiff} & \cite{martin2025pnp} & \cite{xu2025fast} & \cite{tsao2024boosting} &  \\
\midrule
\multirow{4}{*}{\textbf{RealPhoto60}}
& PaQ-2-PiQ$\,\textcolor{red}{\uparrow}$ & 69.04 & 63.38 & 64.40 & 71.88 & 69.97 & 73.77 & \textcolor{blue}{\textbf{75.84}} & 73.41 & 74.42 & 75.27 & 72.78 & 73.82 & 73.44 & \textcolor{red}{\textbf{78.38}} \\

& MANIQA$\,\textcolor{red}{\uparrow}$ & 0.5094 & 0.3759 & 0.4456 & 0.6253 & 0.5290 & 0.6057 & 0.6166 & 0.5989 & 0.6165 & \textcolor{red}{\textbf{0.6527}} & 0.6176 & 0.6348 & 0.6320 & \textcolor{blue}{\textbf{0.6425}} \\

& MUSIQ$\,\textcolor{red}{\uparrow}$ & 59.29 & 45.46 & 57.89 & 63.67 & 64.53 & 70.80 & 70.46 & 70.44 & 70.26 & \textcolor{blue}{\textbf{72.74}} & 70.08 & 71.82 & 70.96 & \textcolor{red}{\textbf{74.81}} \\

& CLIPIQA+$\,\textcolor{red}{\uparrow}$ & 0.4389 & 0.3397 & 0.4214 & 0.4935 & 0.4786 & 0.5691 & 0.5273 & 0.5720 & 0.5528 & \textcolor{blue}{\textbf{0.5932}} & 0.5795 & 0.5842 & 0.5856 & \textcolor{red}{\textbf{0.7861}} \\
\midrule
\multirow{3}{*}{\textbf{RealSR}}
& MANIQA$\,\textcolor{red}{\uparrow}$ & 0.3656 & 0.3660 & 0.3465 & 0.4182 & 0.3855 & 0.5189 & 0.4337 & 0.4052 & 0.4296 & 0.4893 & 0.4910 & 0.5232 & \textcolor{blue}{\textbf{0.5308}} & \textcolor{red}{\textbf{0.6205}} \\

& MUSIQ$\,\textcolor{red}{\uparrow}$ & 62.06 & 64.67 & 61.07 & 61.74 & 62.03 & 69.38 & 65.33 & 64.87 & 66.09 & \textcolor{blue}{\textbf{69.77}} & 62.27 & 68.82 & 63.96 & \textcolor{red}{\textbf{72.34}} \\

& CLIPIQA$\,\textcolor{red}{\uparrow}$ & 0.4872 & 0.5329 & 0.5139 & 0.6202 & 0.5043 & 0.6839 & \textcolor{blue}{\textbf{0.6895}} & 0.5729 & 0.5371 & 0.6804 & 0.6822 & 0.6792 & 0.6842 & \textcolor{red}{\textbf{0.6937}} \\
\midrule
\multirow{3}{*}{\textbf{DrealSR}}
& MANIQA$\,\textcolor{red}{\uparrow}$ & 0.3423 & 0.3420 & 0.3171 & 0.3801 & 0.3456 & 0.4974 & 0.3676 & 0.3572 & 0.4174 & \textcolor{blue}{\textbf{0.5647}} & 0.5228 & 0.5424 & 0.5386 & \textcolor{red}{\textbf{0.6492}} \\

& MUSIQ$\,\textcolor{red}{\uparrow}$ & 58.37 & 61.22 & 56.43 & 55.14 & 53.82 & 67.42 & 59.83 & 66.68 & 64.53 & \textcolor{blue}{\textbf{68.92}} & 67.32 & 68.08 & 68.26 & \textcolor{red}{\textbf{71.82}} \\

& CLIPIQA$\,\textcolor{red}{\uparrow}$ & 0.4847 & 0.5385 & 0.5344 & 0.6005 & 0.5458 & \textcolor{blue}{\textbf{0.7022}} & 0.6620 & 0.6037 & 0.5800 & 0.6848 & 0.6821 & 0.6946 & 0.6884 & \textcolor{red}{\textbf{0.7169}} \\
\bottomrule
\end{tabular}
}
\end{table}

\begin{table}[tb]
\caption{Recognition results on the dataset of Occluded RoadText 2024 \cite{tom2024Icdar} (OCR recognition). The best and second performances are highlighted in \textcolor{red}{\textbf{red}} and \textcolor{blue}{\textbf{blue}}, respectively.}
\label{tab:recognition_results}
\centering
\resizebox{\textwidth}{!}{
\renewcommand{\arraystretch}{0.7}
\footnotesize
\setlength{\tabcolsep}{2pt}
\begin{tabular}{lc ccccccccccc}
\toprule
Metrics & GT & LQ & Real-ESRGAN \cite{wang2021real} & BSRGAN \cite{zhang2021designing} & StableSR \cite{wang2024exploiting} & DiffBIR \cite{lin2024diffbir} & SeeSR \cite{wu2024seesr} & DreamClear \cite{ai2024dreamclear} & SUPIR \cite{yu2024scaling} & FaithDiff \cite{chen2025faithdiff}& Ours \\
\midrule
Precision & 52.72\% & 7.54\% & 13.19\% & 12.04\% & 19.87\% & 26.21\% & 30.07\% & 22.45\% & 31.78\% & \textcolor{blue}{\textbf{36.45\%}} & \textcolor{red}{\textbf{42.23\%}}\\
Recall & 56.67\% & 7.59\% & 13.68\% & 12.33\% & 20.31\% & 27.90\% & 33.09\% & 23.50\% & 41.57\% & \textcolor{blue}{\textbf{46.74\%}} & \textcolor{red}{\textbf{49.75\%}}\\
\bottomrule
\end{tabular}
}
\end{table}

\subsection{Comparison with State-of-the-Art Methods}

We compare our proposed method against several leading approaches, including: GAN-based methods (\textit{i.e.}, Real-ESRGAN \cite{wang2021real} and BSRGAN \cite{zhang2021designing}), Diffusion-based methods (\textit{i.e.}, StableSR \cite{wang2024exploiting}, DiffBIR \cite{lin2024diffbir}, PASD \cite{yang2024pixel}, SeeSR \cite{wu2024seesr}, DreamClear \cite{ai2024dreamclear}, OSEDiff \cite{wu2024one}, SUPIR \cite{yu2024scaling}, and FaithDiff \cite{chen2025faithdiff}) and Flow-based methods (\textit{i.e.}, PnP-Flow \cite{martin2025pnp}, FlowSR \cite{xu2025fast} and LP \cite{tsao2024boosting}). We use PSNR, SSIM, perceptual-oriented metrics (\textit{i.e.}, LPIPS \cite{zhang2018unreasonable}, DISTS \cite{ding2020image}, FID \cite{heusel2017gans}) and no-reference metrics (\textit{i.e.}, MUSIQ \cite{ke2021musiq}, CLIPIQA+ \cite{wang2023exploring}, PaQ-2-PiQ \cite{ying2020patches}, CLIPIQA \cite{wang2023exploring} and MANIQA \cite{yang2022maniqa}) to evaluate the fidelity and quality of restored images. Our method performs strongly across all these metrics, attesting to the high quality of our restorations. Notably, ResFlow-Tuner surpasses all prior methods by a considerable margin in no-reference metrics, establishing a new state-of-the-art. Although resulting in lower PSNR/SSIM scores, recent studies \cite{you2024depicting, yu2024scaling} make compelling cases that these metrics fail to adequately capture perceived visual quality, which is also consistent with our experimental observations.

% \noindent \textbf{Qualitative Comparisons:} Qualitative comparisons are presented in Fig. \ref{fig:3}. In scenarios involving severe degradation (first row), our method stands out as the only approach capable of faithfully inferring the underlying structure while generating crisp details. Competing methods, by contrast, often yield structural distortions or blurry outputs. For real-world images (last three rows), our method consistently produces results with richer detail and a more natural appearance, as supported by more visual comparisons in Appendix.

\subsection{Evaluation on Downstream Benchmarks}
We quantitatively assess our method by evaluating its performance on Optical Character Recognition (OCR) tasks. Following the established evaluation benchmark \cite{chen2025faithdiff}, we conduct experiments on 200 images from the ICDAR 2024 Occluded RoadText dataset \cite{tom2024Icdar}. The recognition is performed with the robust PaddleOCR v3 \cite{li2022pp} framework. As reported in Tab. \ref{tab:recognition_results}, our approach consistently outperforms existing image restoration methods in terms of OCR accuracy. This result provides strong evidence that our method is particularly effective at recovering semantically correct and robust structural information.

\begin{table}[tb]
\caption{Ablation results on LSDIR-Val for ResFlow-Tuner.}
\label{tab:ablation}
\centering
\resizebox{\textwidth}{!}{
\renewcommand{\arraystretch}{0.7}
\footnotesize
\begin{tabular}{@{}lcccccccc@{}}
\toprule
& LPIPS $\textcolor{red}{\downarrow}$ & DISTS $\textcolor{red}{\downarrow}$ & FID $\textcolor{red}{\downarrow}$ & MANIQA $\textcolor{red}{\uparrow}$ & MUSIQ $\textcolor{red}{\uparrow}$ & CLIPIQA $\textcolor{red}{\uparrow}$ & Infer-Compute $\textcolor{red}{\downarrow}$ & Parameters $\textcolor{red}{\downarrow}$ \\
\midrule
LR w/o Text Caption & 0.3856 & 0.1468 & 20.02 & 0.6542 & 72.98 & 0.7212 & / & / \\
LR w/ Text Caption & \textbf{0.3721} & \textbf{0.1406} & \textbf{19.82} & \textbf{0.6658} & \textbf{74.21} & \textbf{0.7389} & / & / \\
\midrule
Direct Adding & 0.4223 & 0.1847 & 28.39 & 0.5796 & 69.14 & 0.6863 & / & 612M / 5.1\% \\
ControlNet & 0.4056 & 0.1663 & 24.73 & 0.6172 & 71.20 & 0.6917 & / & 3.3B / 27.5\% \\
UMMF & \textbf{0.3721} & \textbf{0.1406} & \textbf{19.82} & \textbf{0.6658} & \textbf{74.21} & \textbf{0.7389} & / & \textbf{58.0M / 0.5\%} \\
\midrule
Aesthetic & 0.3872 & 0.1662 & 23.25 & 0.6224 & 72.57 & 0.7128 & / & / \\
CLIPScore & 0.3824 & 0.1524 & 21.04 & 0.5774 & 69.88 & 0.6982 & / & / \\
ImageReward & 0.3789 & 0.1564 & 21.98 & 0.5842 & 70.64 & 0.6874 & / & / \\
\noalign{\vskip 2pt}   % 在虚线上方加 2pt 空隙
\hdashline
\noalign{\vskip 2pt}   % 在虚线上方加 2pt 空隙
Aesthetic + CLIPScore & 0.3820 & 0.1518 & 20.58 & 0.6302 & 73.24 & 0.7181 & / & / \\
Aesthetic + ImageReward & 0.3762 & 0.1558 & 21.76 & 0.6483 & 73.65 & 0.7308 & / & / \\
CLIPScore + ImageReward & 0.3752 & 0.1448 & 20.25 & 0.5964 & 70.98 & 0.7064 & / & / \\
\noalign{\vskip 2pt}   % 在虚线上方加 2pt 空隙
\hdashline
\noalign{\vskip 2pt}   % 在虚线上方加 2pt 空隙
Verifier Ensemble & \textbf{0.3721} & \textbf{0.1406} & \textbf{19.82} & \textbf{0.6658} & \textbf{74.21} & \textbf{0.7389} & / & / \\
\midrule
w/o TTS & 0.4015 & 0.1643 & 22.91 & 0.6032 & 71.47 & 0.6946 & 0.025e4 & / \\
w/ TTS ($K=2, N=5$) & 0.3874 & 0.1523 & 20.87 & 0.6298 & 72.85 & 0.7149 & 0.15e4 & / \\
w/ TTS ($K=4, N=7$) & \textbf{0.3721} & \textbf{0.1406} & \textbf{19.82} & \textbf{0.6658} & \textbf{74.21} & \textbf{0.7389} & \textbf{0.375e4} & / \\
w/ TTS ($K=10, N=15$) & 0.3675 & 0.1332 & 18.17 & 0.6878 & 76.33 & 0.7574 & 1.0e4 & / \\
\bottomrule
\end{tabular}
}
\end{table}

\section{Ablation Study}
\subsection{Impact of Multi-Modal Conditioning}

\noindent \textbf{Effect of Text Caption:} As shown in Tab. \ref{tab:ablation}, incorporating text captions (LR w/ Text Caption) consistently improves performance across all metrics compared to the variant without captions (LR w/o Text Caption). For instance, LPIPS decreases from 0.3856 to 0.3721, DISTS from 0.1468 to 0.1406, and FID from 20.02 to 19.82. Similarly, no-reference quality metrics also exhibit notable gains. These results demonstrate that semantic guidance from text captions effectively steers the generative process toward outputs that are not only pixel-accurate but also perceptually realistic and semantically coherent.

\noindent \textbf{Modality Fusion Strategy:} We compare three fusion strategies: direct feature addition, ControlNet-based conditioning, and our proposed UMMF. Our method achieves the best results across all metrics. For example, it attains an LPIPS of 0.3721, compared to 0.4223 and 0.4056 from the alternatives. Similarly, on no-reference quality metrics such as MANIQA, our method also registers an improvement, outperforming Direct Adding and ControlNet by 14.87\% and 7.87\%, respectively. This superiority stems from our unified sequence modeling formulation, which enables flexible cross-modal token interactions via DiT’s attention mechanism, thereby capturing rich semantic relationships without imposing hard structural constraints.

\noindent \textbf{Efficiency and Cost Analysis:} Unlike traditional "train-time scaling" strategies that rely on heavy adapters (e.g., ControlNet) to inject low-quality image features, our approach unleashes the model prior through a lightweight alignment module. This architectural choice results in a remarkably small number of trainable parameters—only 58M (0.5\%) for the 12B FLUX.1 model. This is an order of magnitude more efficient than adapter-based counterparts. Our parameter-efficient design enables rapid adaptation to new tasks with minimal training overhead, which constitutes a significant efficiency advantage.

\subsection{Study on Verifier Ensemble}

Our TTS framework incorporates three distinct reward verifiers: Aesthetic Score Predictor, CLIPScore, and ImageReward. To validate the necessity of our design choices, we conduct a comprehensive ablation study by either removing or substituting individual components. Specifically, our ablations evaluate (i) the performance of each reward function in isolation and (ii) the performance of all possible pairwise combinations. 

Experimental results indicate that single or pairwise reward models tend to optimize performance within a specific dimension: The Aesthetic Score Predictor is designed to estimate human ratings of synthetic image quality, thereby biasing model optimization toward no-reference evaluation metrics. In contrast, CLIPScore is a reference-free metric derived from the CLIP model. It projects visual and textual embeddings into a shared latent space by leveraging 400M manually annotated image-text pairs, and quantifies semantic consistency via cosine similarity between image and corresponding text embeddings. Consequently, CLIPScore directs optimization towards reference-based metrics that assess text-image alignment. ImageReward, on the other hand, learns broader human preferences through a meticulously designed annotation protocol that encompasses ratings and rankings of text-image alignment, aesthetic quality, and sample harmlessness. 

As shown in Tab. \ref{tab:ablation}, these results demonstrate that only the ensemble of all three reward models, as proposed in this paper, enables a balanced and superior performance across all evaluation dimensions.

\subsection{Analysis of Test-Time Scaling (TTS)}

As summarized in Tab. \ref{tab:ablation}, our method exhibits monotonic performance improvement with increasing inference-time computation (As the number of candidates per round $N$ and intervention rounds $K$ increase), measured by the Number of Function Evaluations (NFEs). For instance, moving from no TTS to the configuration $K=4, N=7$ reduces LPIPS from 0.4015 to 0.3721 and FID from 22.91 to 19.82, while MUSIQ increases from 71.47 to 74.21. A critical observation is that the marginal benefit is not constant. The performance growth rate is highest until the computational cost reaches approximately $15\times$ that of the baseline. This operating point corresponds to a sweet spot on the performance-cost Pareto frontier—beyond this point, the performance gain per unit of additional computation diminishes significantly. Guided by this principle, we set our runtime computation to $15\times$ the baseline ($K=4, N=7$). As mentioned before, when comparing the total cost of our approach (minimal training cost + controlled inference cost) against "train-time scaling" methods (typically high training cost + single inference cost), our total resource consumption is often more favorable. More importantly, our method achieves a level of performance through controlled test-time computation that remains unattainable by others, even with their substantial training investments.

\section{Conclusion}

In this paper, we presented a novel image restoration framework ResFlow-Tuner. Our key innovation lies in the seamless integration of multi-modal guidance and a Test-Time Scaling technique specifically redesigned for ODE-based restoration models. Extensive experiments demonstrate that our method achieves state-of-the-art performance, excelling particularly in generating perceptually realistic and semantically faithful details. This work underscores the immense potential of leveraging large-scale pre-trained flow models for low-level vision tasks and introduces a new direction for post-training optimization of such models. For future work, we plan to explore more efficient TTS strategies to reduce the computational overhead and investigate the application of our framework to other image restoration tasks, such as deblurring and inpainting. The code and models will be made publicly available to facilitate further research.

\section{Details of User Study}
\textbf{Image Selection and Preparation:} To comprehensively validate the perceptual superiority of our proposed method over existing state-of-the-art approaches, we conducted a large-scale, systematic user study. This study moves beyond traditional metrics and focuses on human perceptual preferences, which are the ultimate criterion for perceptual image restoration tasks.
We randomly selected 50 low-quality (LQ) source images from both the synthetic and real-world benchmarks. For each LQ image, we generated its high-quality (HQ) counterpart using six different methods: Ours, and five competing state-of-the-art approaches (\textit{i.e}., Real-ESRGAN, DiffBIR, SeeSR, SUPIR, and FaithDiff). This resulted in a total of $50 \times 6 = 300$ restored images for evaluation.

\noindent \textbf{Participants:} We recruited 64 participants (30 females, 34 males, aged between 18 and 45) for the study. All participants were screened for normal or corrected-to-normal visual acuity. The sample size was determined to ensure statistical power and reliability, exceeding the common minimum in usability studies  and allowing for robust statistical analysis across multiple image sets. Participants were compensated for their time.

\noindent \textbf{Assessment Methodology:} Given the fine-grained nature of comparing outputs from modern IR algorithms, we employed a \textbf{forced-choice pairwise comparison} methodology, which is more sensitive and reliable for discerning subtle perceptual differences than absolute rating scale. For each of the 50 LQ source images, participants were presented with a series of side-by-side image pairs, each pair displaying the restored results from two randomly selected methods (from the pool of six). The presentation order of methods and image pairs was fully randomized and balanced across participants to eliminate ordering biases. On each trial, participants were instructed to select the image in the pair that they perceived as having higher overall visual quality, considering the factors of Visual Quality (sharpness, clarity), Naturalness (freedom from artifacts, realistic texture), and Accuracy (faithfulness to the expected content and structure of the original scene). They were forced to make a choice; "no preference" options were excluded to maximize discriminatory power. Each unique method pair for each source image was compared an equal number of times. With 6 methods, there are $\binom{6}{2} = 15$ unique pairs per image. The total number of comparisons was managed across participants to ensure comprehensive data collection.

\noindent \textbf{Ranking Calculation:} We adopted the \textbf{Bradley-Terry model} to convert the pairwise comparison wins into a continuous preference score for each method. The model estimates the probability $P_{i>j}$ that method $i$ is preferred over method $j$ as:
    \begin{equation}
        P_{i>j}=\frac{\pi_i}{\pi_i+\pi_j}
    \end{equation}
where $\pi_i$ and $\pi_j$ are the positive-valued preference scores for methods $i$ and $j$, respectively. The scores $\pi$ for all methods are estimated from the observed comparison data using maximum likelihood estimation. The final global rank is determined by sorting these $\pi$ scores in descending order.

\noindent \textbf{Top-K Ratio:} The frequency with which a method was ranked among the top-K selections across all image groups, defined as:
    \begin{equation}
        R_i^k = \frac{1}{N} \sum_{j=1}^N \mathbb{I}(i \in F_{\text{topk}}(s_j, k))
    \end{equation}
where:

\begin{itemize}
\item $N = 100$ is the total number of image groups
\item $s_j = \{s_{ij} \mid i = 1, \ldots, 6\}$ represents the selection scores for the $j$-th image group
\item $F_{\text{topk}}$ denotes the top-K ranking operation
\item $\mathbb{I}(\cdot)$ is the indicator function
\end{itemize}

This metric captures both the proportion of most preferred images and the consistency in producing high-quality results across diverse content.

\begin{figure*}[htbp]
    \centering
    \includegraphics[width=\linewidth]{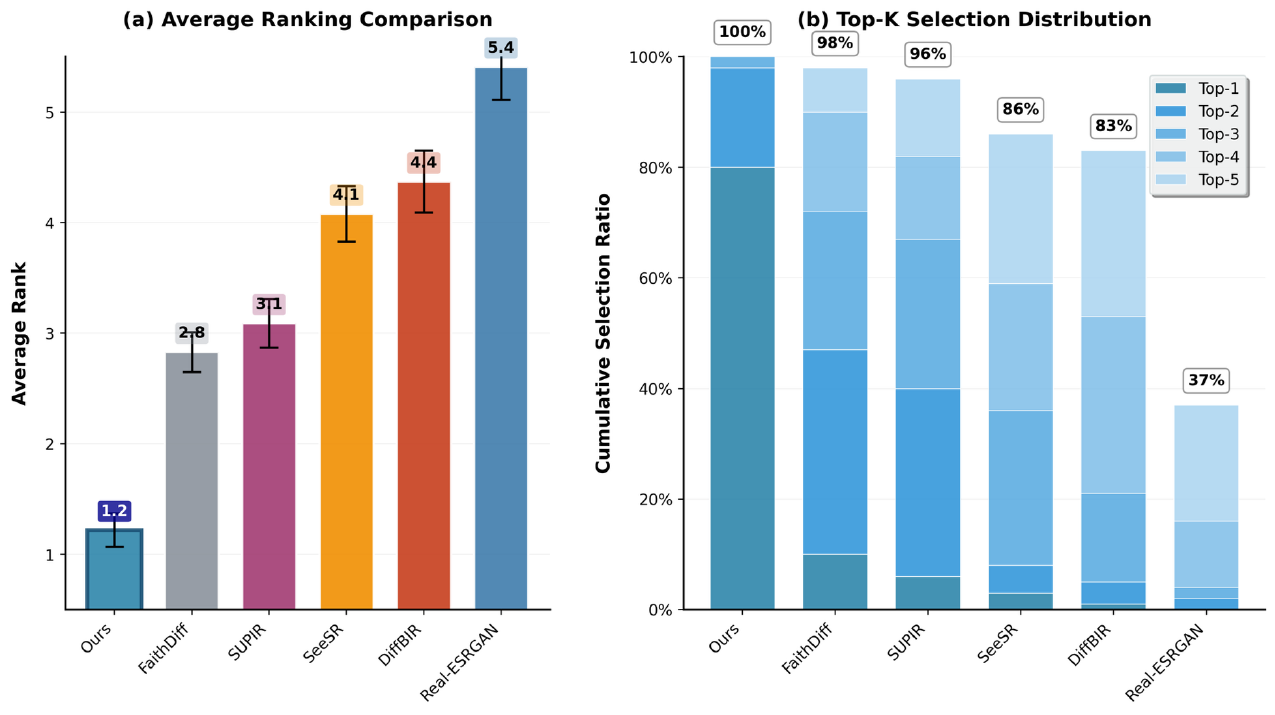}
    \caption{User Study Results. (a) Average ranking of the six methods across all participants and test images, with error bars. (b) Top-K ratios (K=1,2,3,4,5) demonstrating our method's consistency in producing high-quality results across diverse image content.}
    \label{fig:user study}
\end{figure*}

\noindent \textbf{Results and Visualization:} The results of our large-scale user study provide compelling evidence for the superiority of our proposed method.
\begin{itemize}
    \item \textbf{Overall Ranking:} The calculated average ranks (based on the Bradley-Terry scores) clearly demonstrate that our method achieved the highest average rank, significantly outperforming all five competing methods. A visualization of the average ranks is presented in Fig. \ref{fig:user study}(a).
    \item \textbf{The Top-K Analysis:} Our method was selected as the single best restoration for 80\% of images, demonstrating its ability to consistently produce the most preferred results. Additionally, our approach ranked within the top two choices for 98\% of images, highlighting exceptional reliability and user satisfaction across diverse image content. as illustrated in the Critical Difference diagram in Fig. \ref{fig:user study}(b).
\end{itemize}

\section{Reward Model Selection}
Given the inherent complexity of image restoration, a comprehensive and fine-grained evaluation is essential. To this end, we complement our evaluation with a suite of supervised verifiers to assess a broader range of image aspects: the Aesthetic Score Predictor \cite{schuhmann2022laion}, CLIPScore \cite{hessel2021clipscore}, and ImageReward \cite{xu2023imagereward}. Trained on extensive human-annotated data, these models capture diverse dimensions of human preference.
\begin{enumerate}
    \item \textbf{Aesthetic Score Predictor} is trained to predict the human rating of synthesized images’ visual quality, often independent of text prompts.
    \item \textbf{CLIPScore} is a reference-free evaluation metric derived from the CLIP model \, which aligns visual and textual embeddings in a shared latent space via 400M human labeled (image, text) pair data. By computing the cosine similarity between an image embedding and its associated text prompt embedding, CLIPScore quantifies semantic coherence without requiring ground-truth images.
    \item \textbf{ImageReward} is a text-to-image human preference reward model [85], which takes an image and its corresponding prompt as inputs and outputs a preference score, including rating and ranking samples on text-image alignment, aesthetic quality, and harmlessness.
\end{enumerate}
To create a more robust evaluator, we integrate the three verifiers into a unified Verifier Ensemble. Since these verifiers operate on different scales, we leverage their relative rankings rather than absolute scores. Specifically, we first record the ordinal ranking of each sample per metric, then aggregate the three rankings via an unweighted average for each sample. The candidate with the highest mean rank is selected.

\section{Visual results for Ablation}
\begin{figure*}[ht]
    \centering
    
    % 第一行图片 - 使用替代\hfill来减小列间距
    \begin{subfigure}[b]{0.3\textwidth}
        \centering
        \includegraphics[width=0.95\linewidth]{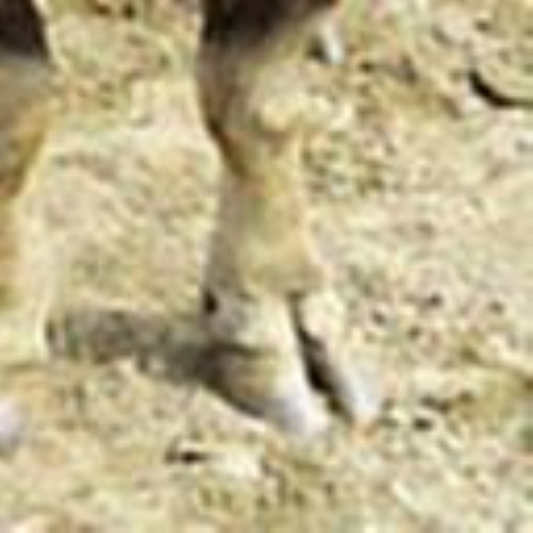}
    \end{subfigure}%
    % 减小列间距
    \begin{subfigure}[b]{0.3\textwidth}
        \centering
        \includegraphics[width=0.95\linewidth]{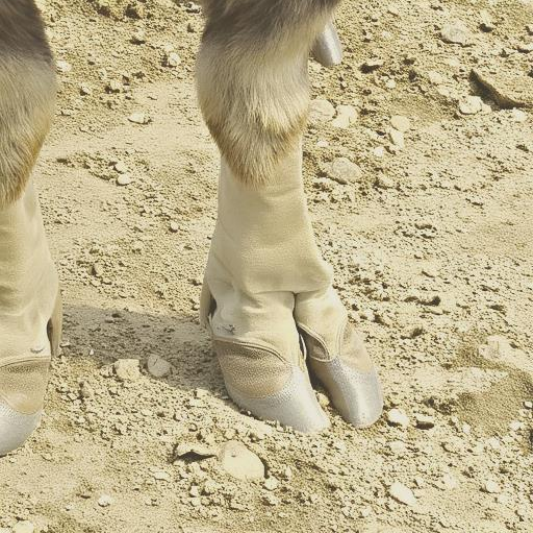}
    \end{subfigure}%
    % 减小列间距
    \begin{subfigure}[b]{0.3\textwidth}
        \centering
        \includegraphics[width=0.95\linewidth]{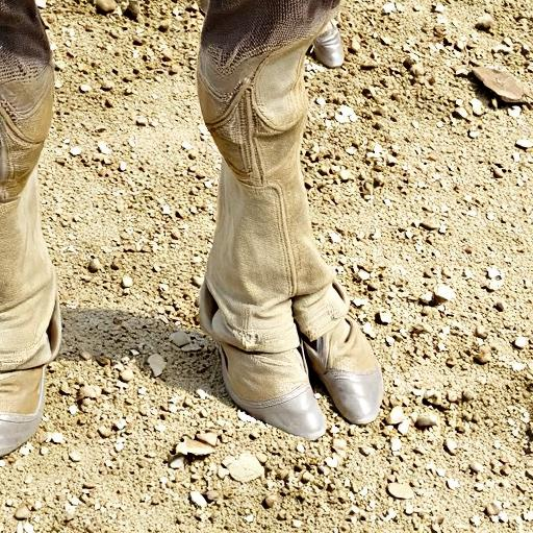}
    \end{subfigure}
    
    % 第一行方法标注
    \vspace{2pt}
    \noindent
    \begin{minipage}[t]{0.3\textwidth}
        \centering LQ Input
    \end{minipage}%
    % 同样减小标注的列间距
    \begin{minipage}[t]{0.3\textwidth}
        \centering w/o TTS
    \end{minipage}%
    % 同样减小标注的列间距
    \begin{minipage}[t]{0.3\textwidth}
        \centering w/o Text Prompt, w/ TTS
    \end{minipage}

    % 第二行图片
    \begin{subfigure}[b]{0.3\textwidth}
        \centering
        \includegraphics[width=0.95\linewidth]{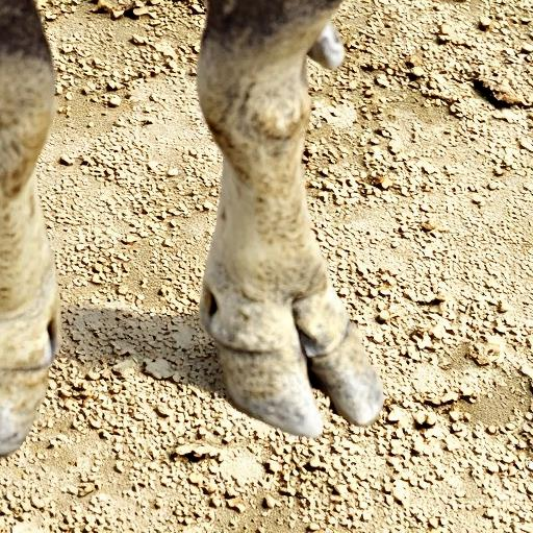}
    \end{subfigure}%
    \begin{subfigure}[b]{0.3\textwidth}
        \centering
        \includegraphics[width=0.95\linewidth]{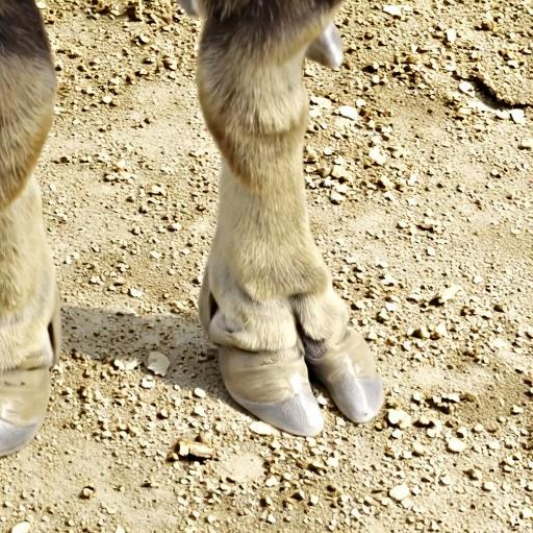}
    \end{subfigure}%
    \begin{subfigure}[b]{0.3\textwidth}
        \centering
        \includegraphics[width=0.95\linewidth]{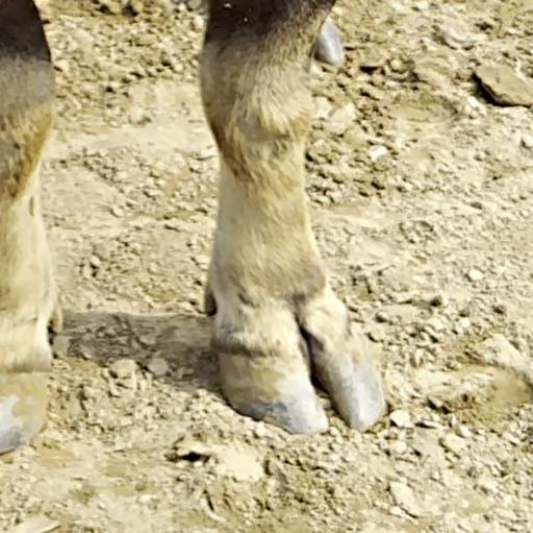}
    \end{subfigure}
    
    % 第二行方法标注
    \vspace{2pt}
    \noindent
    \begin{minipage}[t]{0.3\textwidth}
        \centering w/o Reward Ensemble, w/ TTS
    \end{minipage}%
    \begin{minipage}[t]{0.3\textwidth}
        \centering ResFlow-Tuner (Ours)
    \end{minipage}%
    \begin{minipage}[t]{0.3\textwidth}
        \centering GT
    \end{minipage}
    
    \vspace{0.3cm}
    
    % 第三行图片
    \begin{subfigure}[b]{0.3\textwidth}
        \centering
        \includegraphics[width=0.95\linewidth]{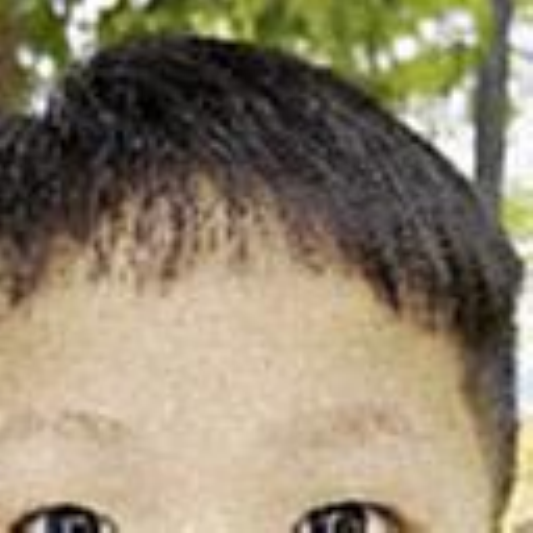}
    \end{subfigure}%
    \begin{subfigure}[b]{0.3\textwidth}
        \centering
        \includegraphics[width=0.95\linewidth]{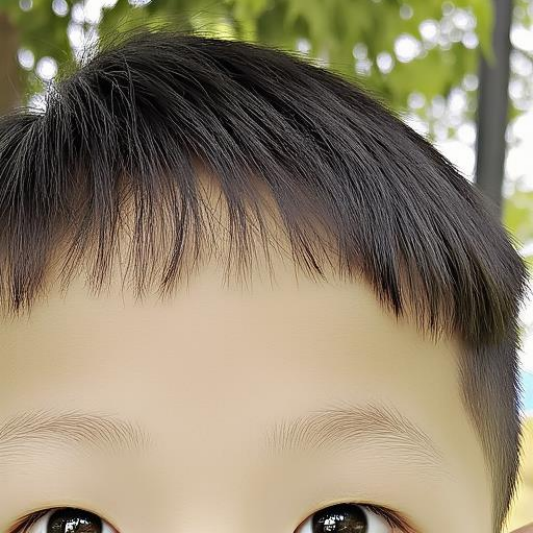}
    \end{subfigure}%
    \begin{subfigure}[b]{0.3\textwidth}
        \centering
        \includegraphics[width=0.95\linewidth]{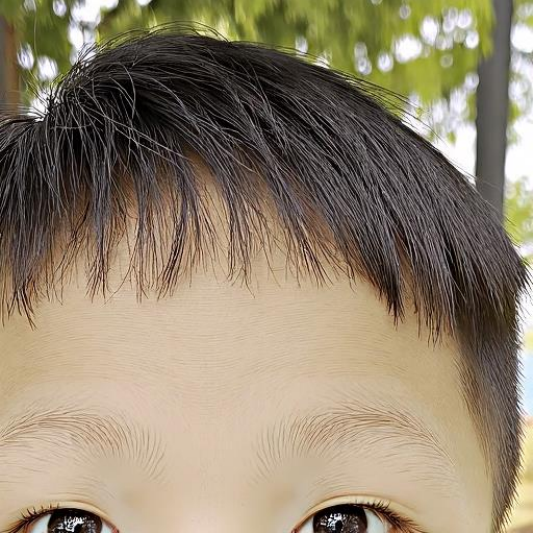}
    \end{subfigure}
    
    % 第三行方法标注
    \vspace{2pt}
    \noindent
    \begin{minipage}[t]{0.3\textwidth}
        \centering LQ Input
    \end{minipage}%
    % 同样减小标注的列间距
    \begin{minipage}[t]{0.3\textwidth}
        \centering w/o TTS
    \end{minipage}%
    % 同样减小标注的列间距
    \begin{minipage}[t]{0.3\textwidth}
        \centering w/o Text Prompt, w/ TTS
    \end{minipage}

    % 第四行图片
    \begin{subfigure}[b]{0.3\textwidth}
        \centering
        \includegraphics[width=0.95\linewidth]{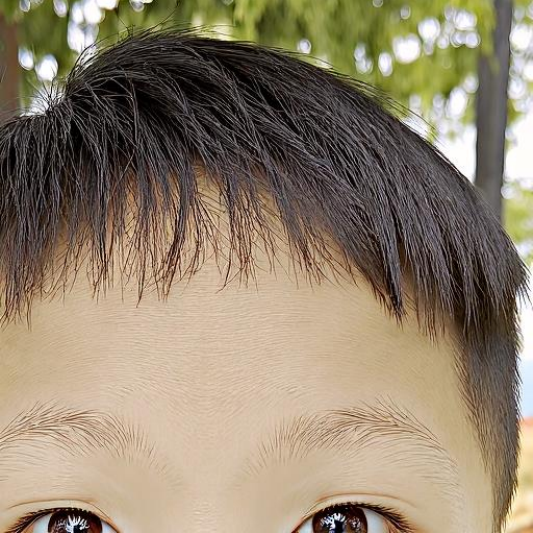}
    \end{subfigure}%
    \begin{subfigure}[b]{0.3\textwidth}
        \centering
        \includegraphics[width=0.95\linewidth]{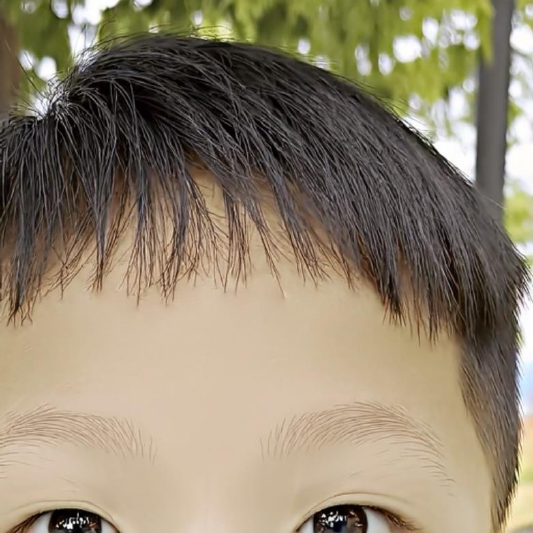}
    \end{subfigure}%
    \begin{subfigure}[b]{0.3\textwidth}
        \centering
        \includegraphics[width=0.95\linewidth]{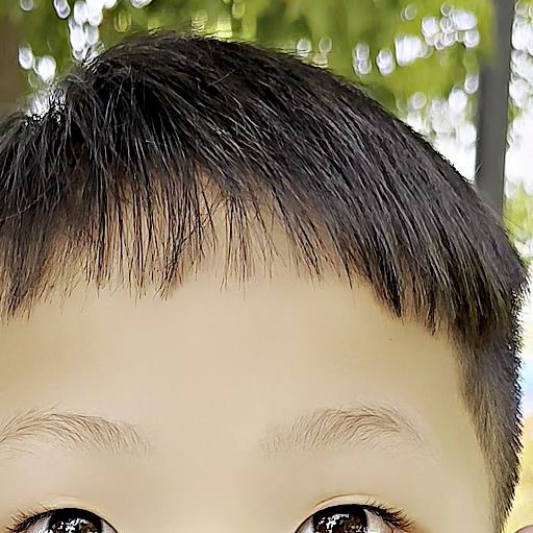}
    \end{subfigure}
    
    % 第四行方法标注
    \vspace{2pt}
    \noindent
    \begin{minipage}[t]{0.3\textwidth}
        \centering w/o Reward Ensemble, w/ TTS
    \end{minipage}%
    \begin{minipage}[t]{0.3\textwidth}
        \centering ResFlow-Tuner (Ours)
    \end{minipage}%
    \begin{minipage}[t]{0.3\textwidth}
        \centering GT
    \end{minipage}
    
    \caption{Visual comparisons for ablation study on ResFlow-Tuner (1/2).}
    \label{fig:ablation-1}
\end{figure*}

\begin{figure*}[ht]
    \centering
    
    % 第一行图片 - 使用替代\hfill来减小列间距
    \begin{subfigure}[b]{0.3\textwidth}
        \centering
        \includegraphics[width=0.95\linewidth]{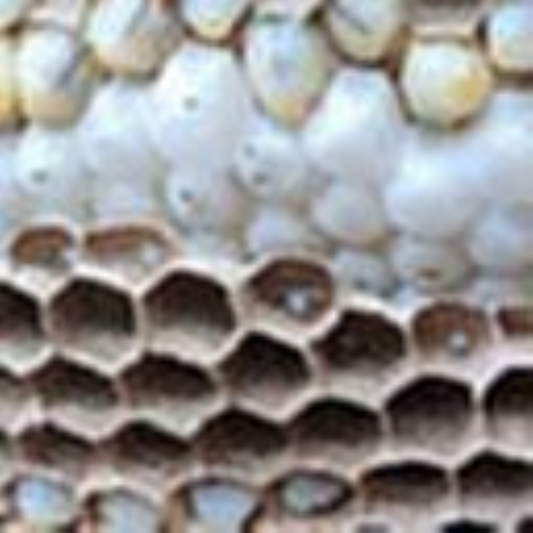}
    \end{subfigure}%
    % 减小列间距
    \begin{subfigure}[b]{0.3\textwidth}
        \centering
        \includegraphics[width=0.95\linewidth]{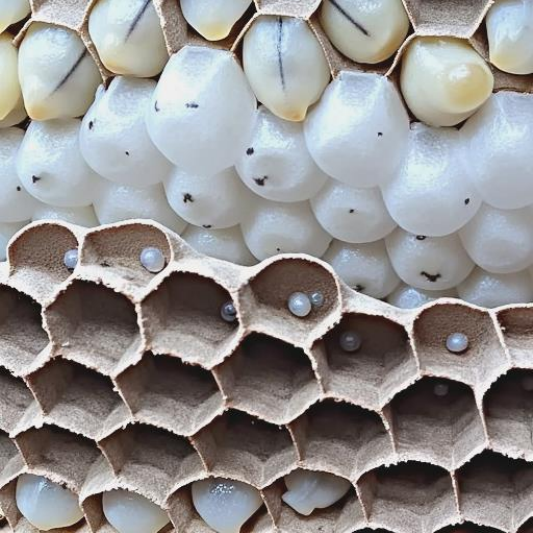}
    \end{subfigure}%
    % 减小列间距
    \begin{subfigure}[b]{0.3\textwidth}
        \centering
        \includegraphics[width=0.95\linewidth]{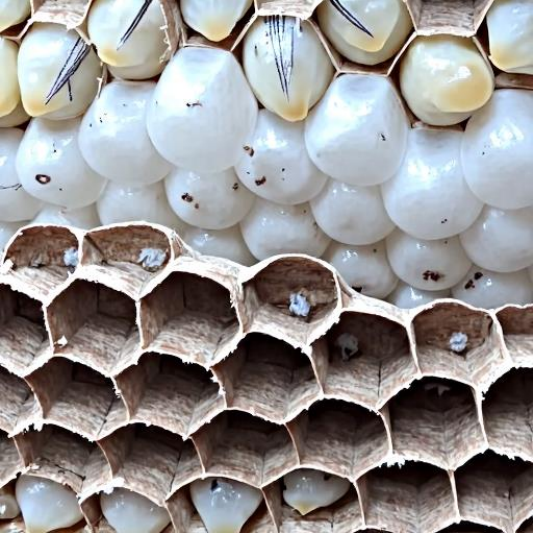}
    \end{subfigure}
    
    % 第一行方法标注
    \vspace{2pt}
    \noindent
    \begin{minipage}[t]{0.3\textwidth}
        \centering LQ Input
    \end{minipage}%
    % 同样减小标注的列间距
    \begin{minipage}[t]{0.3\textwidth}
        \centering w/o TTS
    \end{minipage}%
    % 同样减小标注的列间距
    \begin{minipage}[t]{0.3\textwidth}
        \centering w/o Text Prompt, w/ TTS
    \end{minipage}

    % 第二行图片
    \begin{subfigure}[b]{0.3\textwidth}
        \centering
        \includegraphics[width=0.95\linewidth]{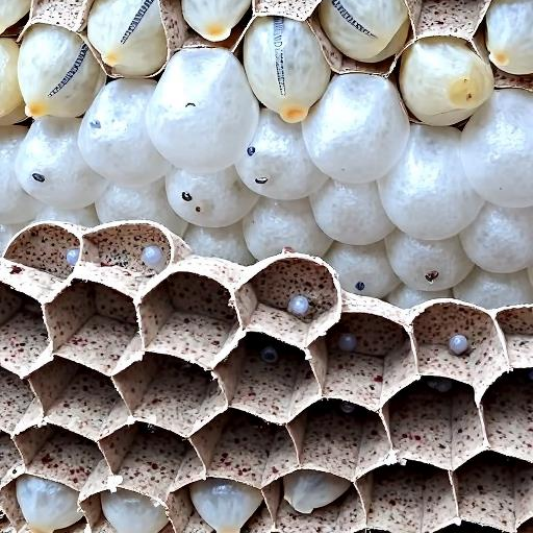}
    \end{subfigure}%
    \begin{subfigure}[b]{0.3\textwidth}
        \centering
        \includegraphics[width=0.95\linewidth]{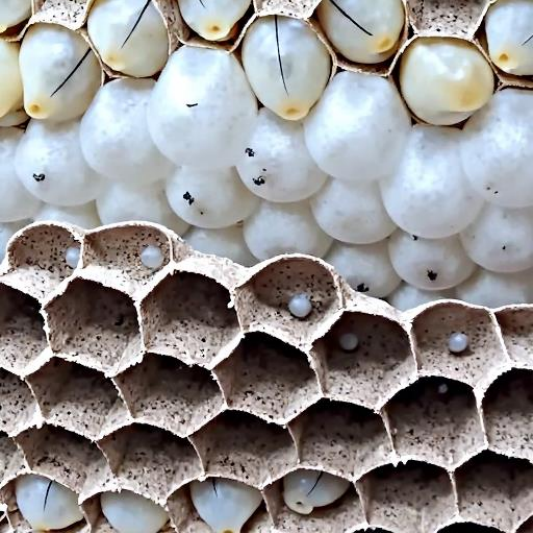}
    \end{subfigure}%
    \begin{subfigure}[b]{0.3\textwidth}
        \centering
        \includegraphics[width=0.95\linewidth]{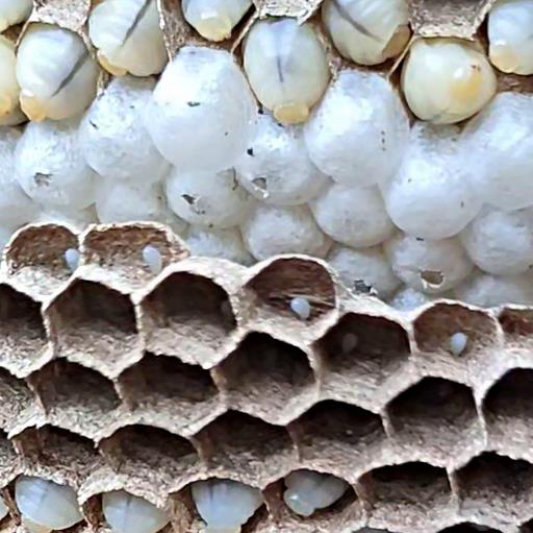}
    \end{subfigure}
    
    % 第二行方法标注
    \vspace{2pt}
    \noindent
    \begin{minipage}[t]{0.3\textwidth}
        \centering w/o Reward Ensemble, w/ TTS
    \end{minipage}%
    \begin{minipage}[t]{0.3\textwidth}
        \centering ResFlow-Tuner (Ours)
    \end{minipage}%
    \begin{minipage}[t]{0.3\textwidth}
        \centering GT
    \end{minipage}
    
    \vspace{0.3cm}
    
    % 第三行图片
    \begin{subfigure}[b]{0.3\textwidth}
        \centering
        \includegraphics[width=0.95\linewidth]{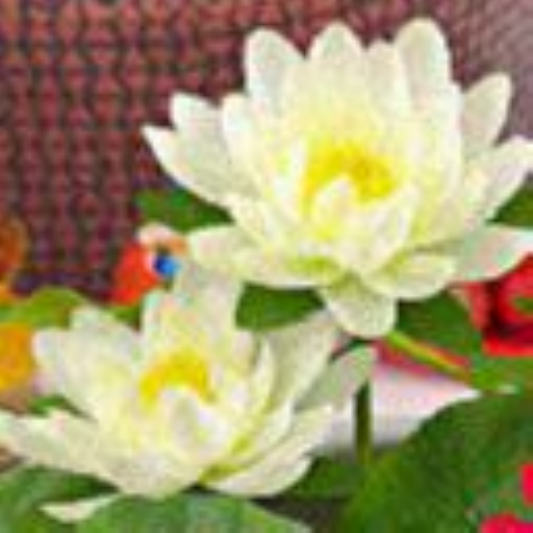}
    \end{subfigure}%
    \begin{subfigure}[b]{0.3\textwidth}
        \centering
        \includegraphics[width=0.95\linewidth]{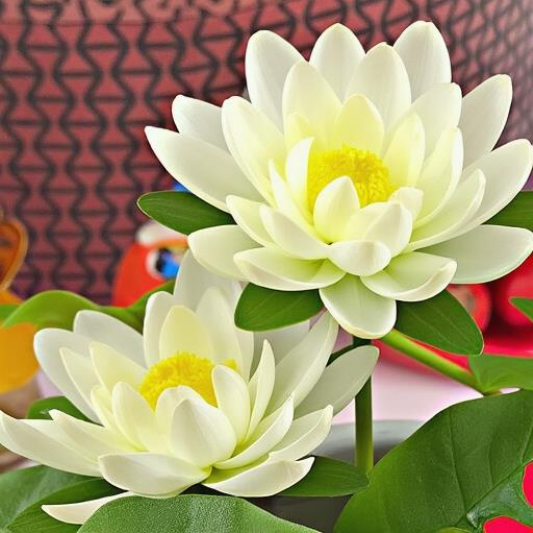}
    \end{subfigure}%
    \begin{subfigure}[b]{0.3\textwidth}
        \centering
        \includegraphics[width=0.95\linewidth]{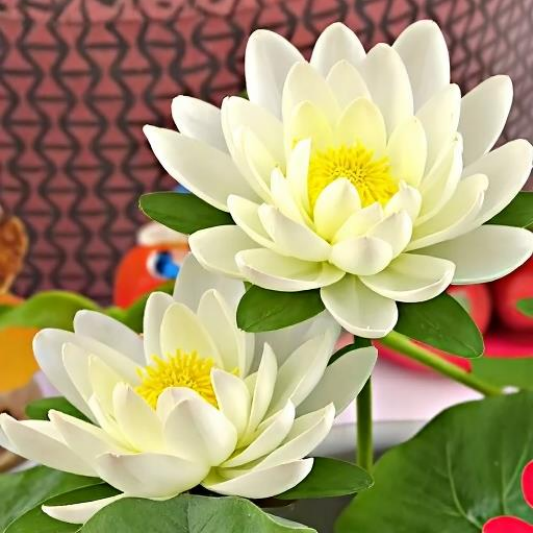}
    \end{subfigure}
    
    % 第三行方法标注
    \vspace{2pt}
    \noindent
    \begin{minipage}[t]{0.3\textwidth}
        \centering LQ Input
    \end{minipage}%
    % 同样减小标注的列间距
    \begin{minipage}[t]{0.3\textwidth}
        \centering w/o TTS
    \end{minipage}%
    % 同样减小标注的列间距
    \begin{minipage}[t]{0.3\textwidth}
        \centering w/o Text Prompt, w/ TTS
    \end{minipage}

    % 第四行图片
    \begin{subfigure}[b]{0.3\textwidth}
        \centering
        \includegraphics[width=0.95\linewidth]{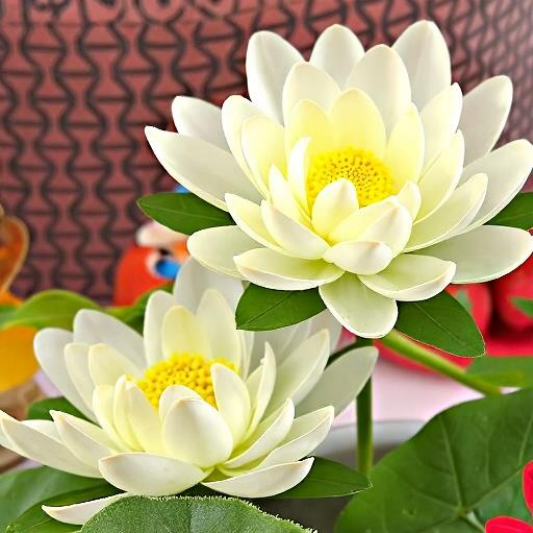}
    \end{subfigure}%
    \begin{subfigure}[b]{0.3\textwidth}
        \centering
        \includegraphics[width=0.95\linewidth]{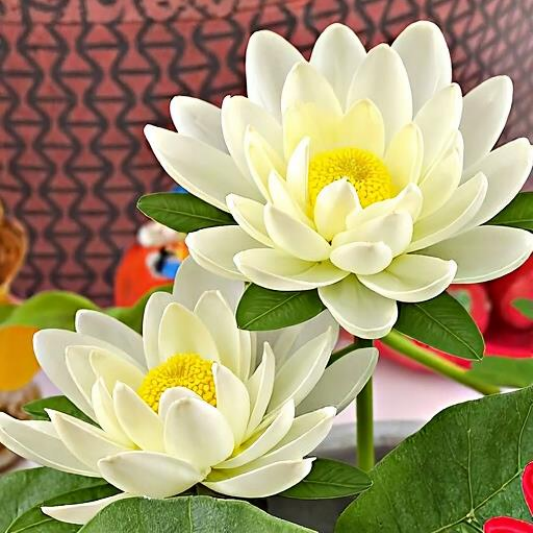}
    \end{subfigure}%
    \begin{subfigure}[b]{0.3\textwidth}
        \centering
        \includegraphics[width=0.95\linewidth]{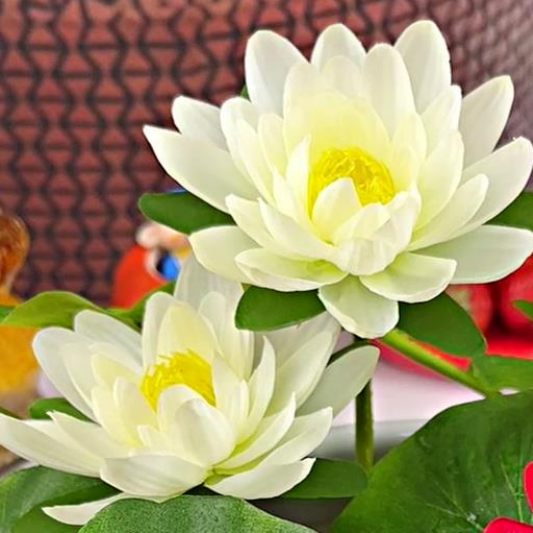}
    \end{subfigure}
    
    % 第四行方法标注
    \vspace{2pt}
    \noindent
    \begin{minipage}[t]{0.3\textwidth}
        \centering w/o Reward Ensemble, w/ TTS
    \end{minipage}%
    \begin{minipage}[t]{0.3\textwidth}
        \centering ResFlow-Tuner (Ours)
    \end{minipage}%
    \begin{minipage}[t]{0.3\textwidth}
        \centering GT
    \end{minipage}
    
    \caption{Visual comparisons for ablation study on ResFlow-Tuner (2/2).}
    \label{fig:ablation-2}
\end{figure*}

We provide more visual comparison results in Fig. \ref{fig:ablation-1} and Fig. \ref{fig:ablation-2}. We find that when using a null prompt instead of a text prompt generated by MLLM, there are significant semantic errors in the restoration results. This demonstrates that the semantic information provided by MLLM-generated detailed text prompts helps the model achieve more ideal restoration results. In contrast, using the original FLUX model alone for conditional restoration (without the subsequent TTS component) results in a significant deterioration of the restoration outcomes, underscoring the necessity of our full pipeline. Similarly, incorporating our reward ensemble into the pipeline leads to notably enhanced texture detail and a markedly more realistic visual appearance.

\section{More Visual Comparisons}
\begin{figure*}[ht]
    \centering
    
    % 第一行图片 - 使用替代\hfill来减小列间距
    \begin{subfigure}[b]{0.3\textwidth}
        \centering
        \includegraphics[width=0.95\linewidth]{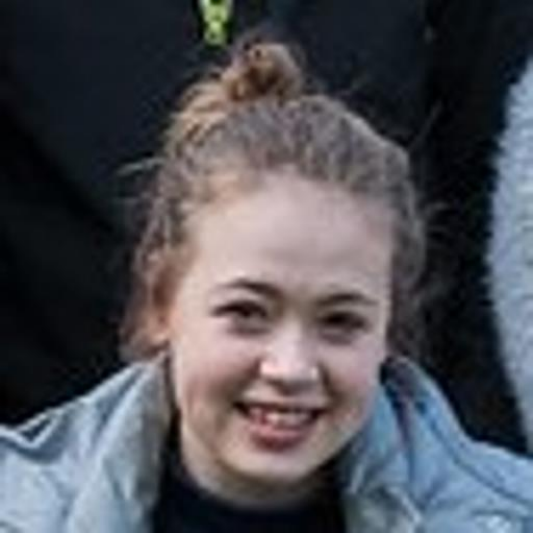}
    \end{subfigure}%
    % 减小列间距
    \begin{subfigure}[b]{0.3\textwidth}
        \centering
        \includegraphics[width=0.95\linewidth]{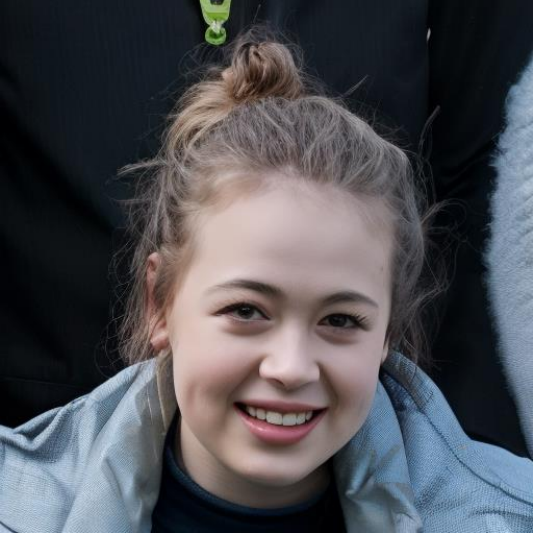}
    \end{subfigure}%
    % 减小列间距
    \begin{subfigure}[b]{0.3\textwidth}
        \centering
        \includegraphics[width=0.95\linewidth]{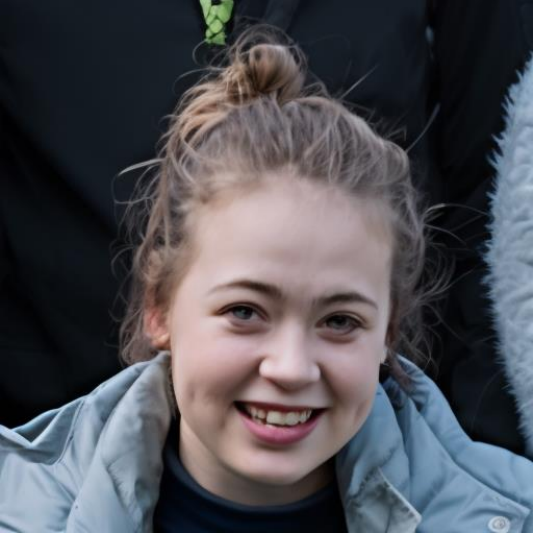}
    \end{subfigure}
    
    % 第一行方法标注
    \vspace{2pt}
    \noindent
    \begin{minipage}[t]{0.3\textwidth}
        \centering LQ Input
    \end{minipage}%
    % 同样减小标注的列间距
    \begin{minipage}[t]{0.3\textwidth}
        \centering DiffBIR \cite{lin2024diffbir}
    \end{minipage}%
    % 同样减小标注的列间距
    \begin{minipage}[t]{0.3\textwidth}
        \centering SeeSR \cite{wu2024seesr}
    \end{minipage}

    % 第二行图片
    \begin{subfigure}[b]{0.3\textwidth}
        \centering
        \includegraphics[width=0.95\linewidth]{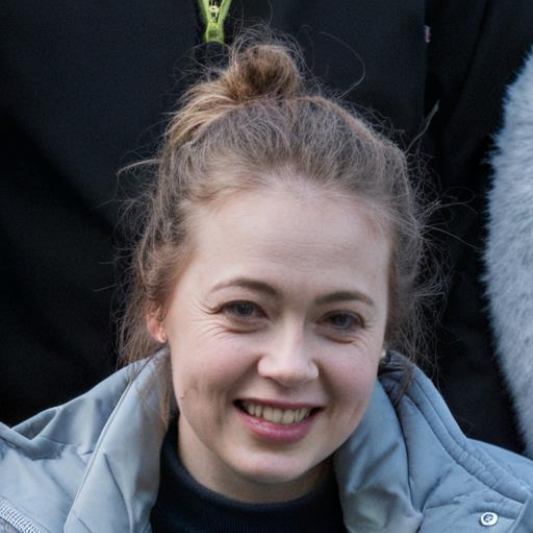}
    \end{subfigure}%
    \begin{subfigure}[b]{0.3\textwidth}
        \centering
        \includegraphics[width=0.95\linewidth]{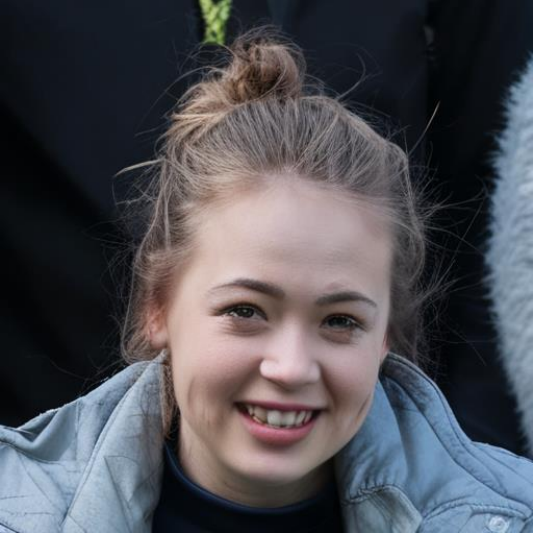}
    \end{subfigure}%
    \begin{subfigure}[b]{0.3\textwidth}
        \centering
        \includegraphics[width=0.95\linewidth]{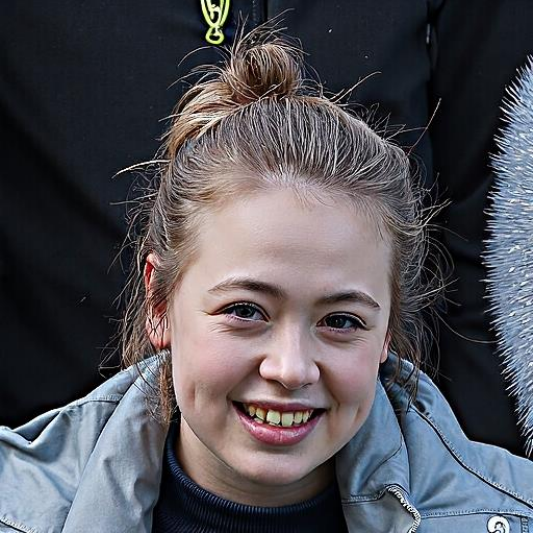}
    \end{subfigure}
    
    % 第二行方法标注
    \vspace{2pt}
    \noindent
    \begin{minipage}[t]{0.3\textwidth}
        \centering SUPIR \cite{yu2024scaling}
    \end{minipage}%
    \begin{minipage}[t]{0.3\textwidth}
        \centering FaithDiff \cite{chen2025faithdiff}
    \end{minipage}%
    \begin{minipage}[t]{0.3\textwidth}
        \centering \textbf{Ours}
    \end{minipage}
    
    \vspace{0.3cm}
    
    % 第三行图片
    \begin{subfigure}[b]{0.3\textwidth}
        \centering
        \includegraphics[width=0.95\linewidth]{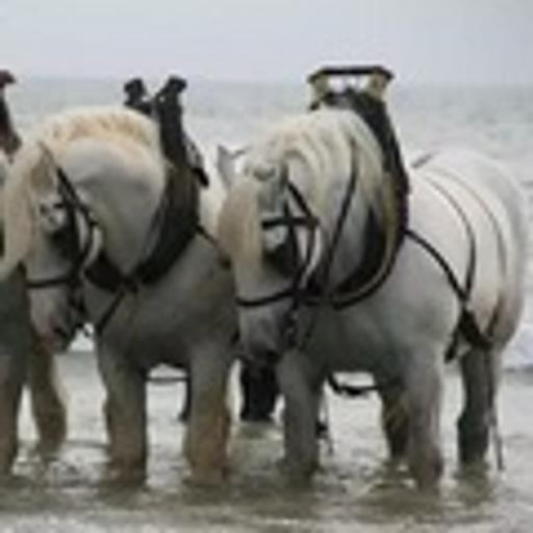}
    \end{subfigure}%
    \begin{subfigure}[b]{0.3\textwidth}
        \centering
        \includegraphics[width=0.95\linewidth]{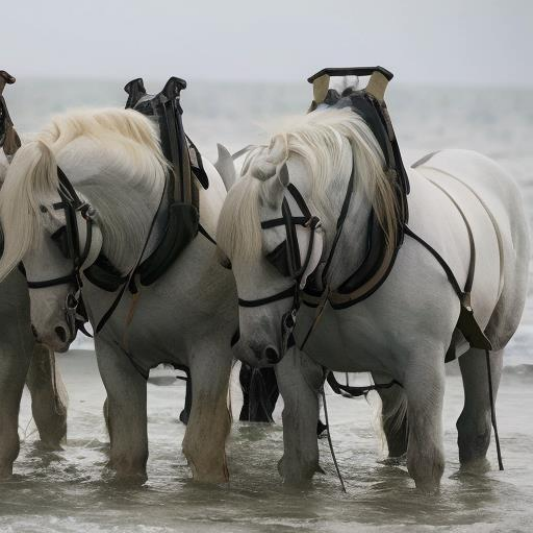}
    \end{subfigure}%
    \begin{subfigure}[b]{0.3\textwidth}
        \centering
        \includegraphics[width=0.95\linewidth]{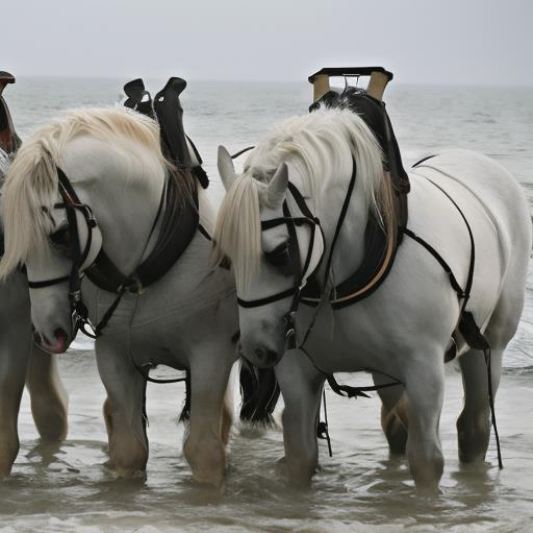}
    \end{subfigure}
    
    % 第三行方法标注
    \vspace{2pt}
    \noindent
    \begin{minipage}[t]{0.3\textwidth}
        \centering LQ Input
    \end{minipage}%
    % 同样减小标注的列间距
    \begin{minipage}[t]{0.3\textwidth}
        \centering DiffBIR \cite{lin2024diffbir}
    \end{minipage}%
    % 同样减小标注的列间距
    \begin{minipage}[t]{0.3\textwidth}
        \centering SeeSR \cite{wu2024seesr}
    \end{minipage}

    % 第四行图片
    \begin{subfigure}[b]{0.3\textwidth}
        \centering
        \includegraphics[width=0.95\linewidth]{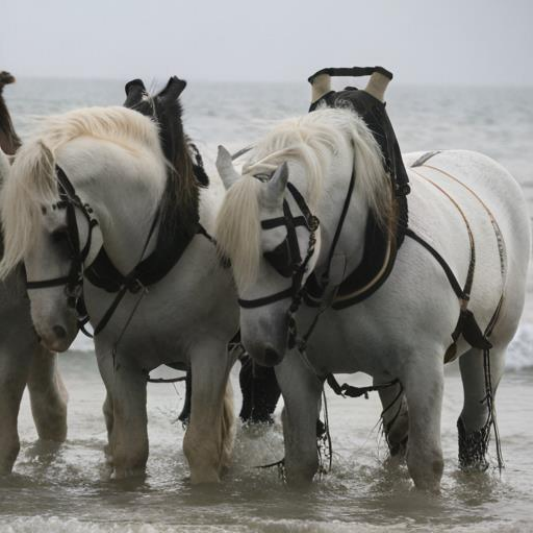}
    \end{subfigure}%
    \begin{subfigure}[b]{0.3\textwidth}
        \centering
        \includegraphics[width=0.95\linewidth]{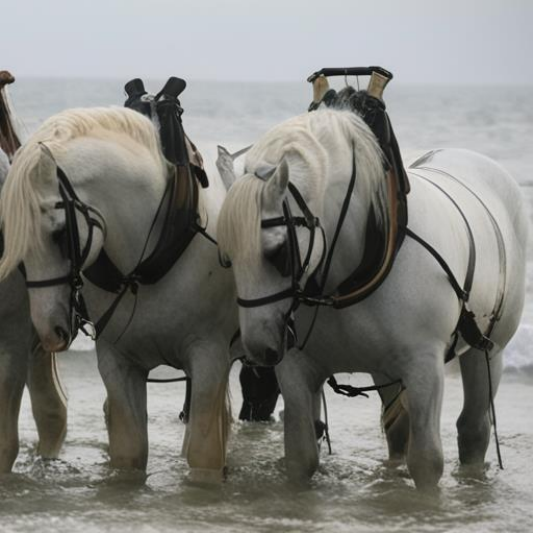}
    \end{subfigure}%
    \begin{subfigure}[b]{0.3\textwidth}
        \centering
        \includegraphics[width=0.95\linewidth]{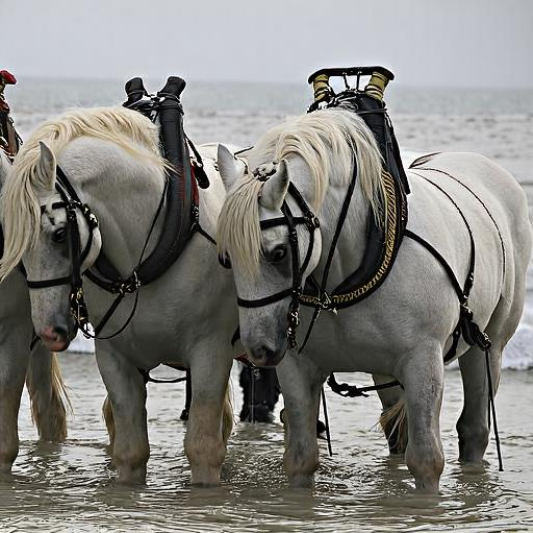}
    \end{subfigure}
    
    % 第四行方法标注
    \vspace{2pt}
    \noindent
    \begin{minipage}[t]{0.3\textwidth}
        \centering SUPIR \cite{yu2024scaling}
    \end{minipage}%
    \begin{minipage}[t]{0.3\textwidth}
        \centering FaithDiff \cite{chen2025faithdiff}
    \end{minipage}%
    \begin{minipage}[t]{0.3\textwidth}
        \centering \textbf{Ours}
    \end{minipage}
    
    \caption{Visual comparisons on real-world benchmarks (1/2). Please zoom in for a better view.}
    \label{fig:real-world-1}
\end{figure*}

\begin{figure*}[ht]
    \centering
    
    % 第一行图片 - 使用替代\hfill来减小列间距
    \begin{subfigure}[b]{0.3\textwidth}
        \centering
        \includegraphics[width=0.95\linewidth]{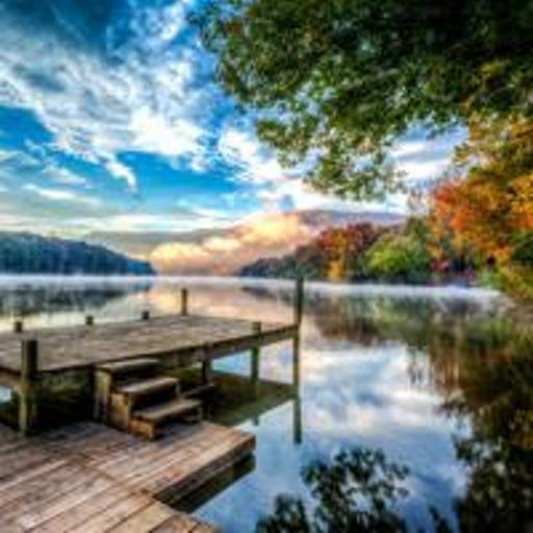}
    \end{subfigure}%
    % 减小列间距
    \begin{subfigure}[b]{0.3\textwidth}
        \centering
        \includegraphics[width=0.95\linewidth]{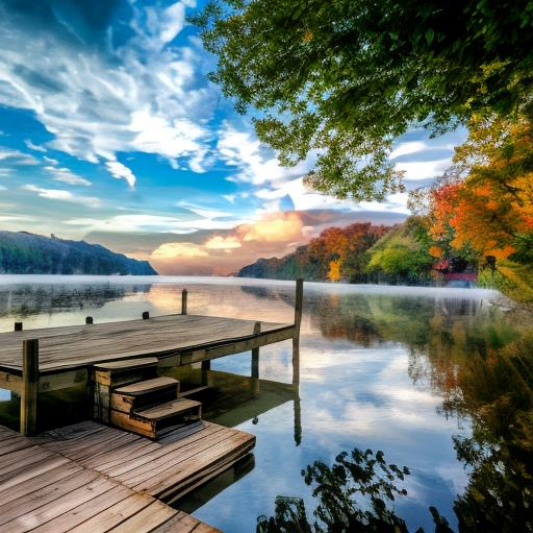}
    \end{subfigure}%
    % 减小列间距
    \begin{subfigure}[b]{0.3\textwidth}
        \centering
        \includegraphics[width=0.95\linewidth]{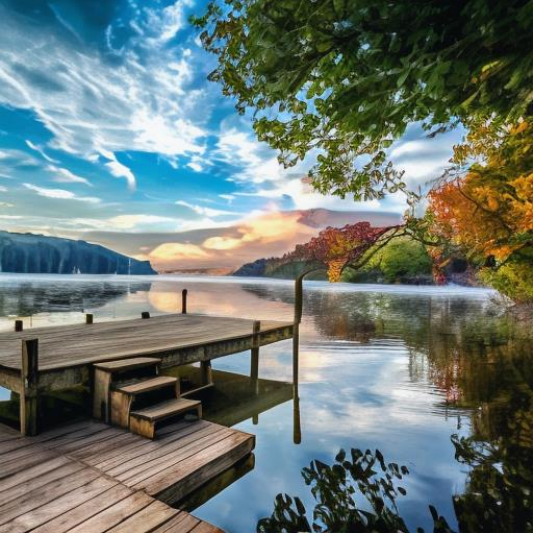}
    \end{subfigure}
    
    % 第一行方法标注
    \vspace{2pt}
    \noindent
    \begin{minipage}[t]{0.3\textwidth}
        \centering LQ Input
    \end{minipage}%
    % 同样减小标注的列间距
    \begin{minipage}[t]{0.3\textwidth}
        \centering DiffBIR \cite{lin2024diffbir}
    \end{minipage}%
    % 同样减小标注的列间距
    \begin{minipage}[t]{0.3\textwidth}
        \centering SeeSR \cite{wu2024seesr}
    \end{minipage}

    % 第二行图片
    \begin{subfigure}[b]{0.3\textwidth}
        \centering
        \includegraphics[width=0.95\linewidth]{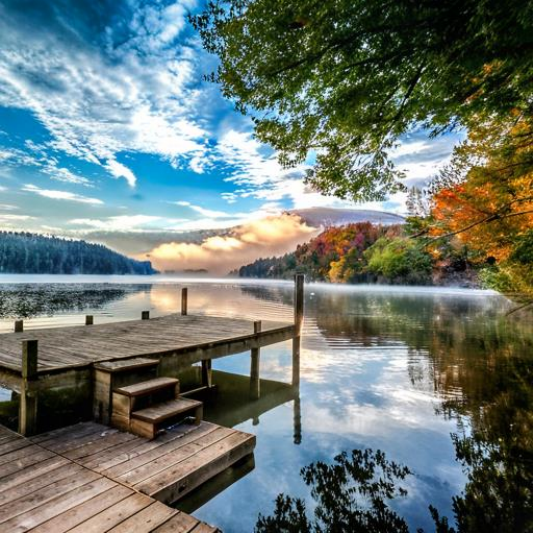}
    \end{subfigure}%
    \begin{subfigure}[b]{0.3\textwidth}
        \centering
        \includegraphics[width=0.95\linewidth]{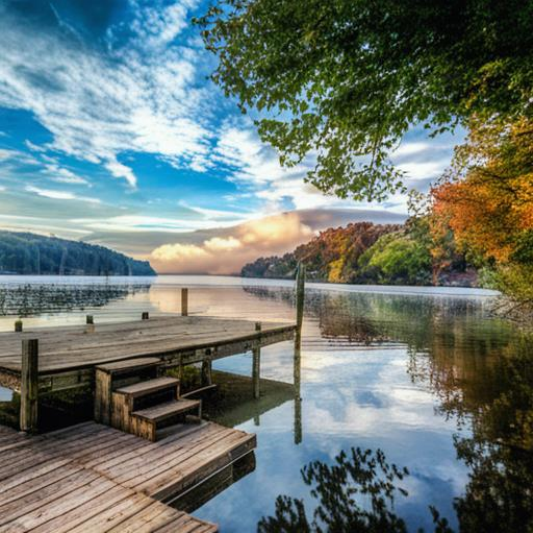}
    \end{subfigure}%
    \begin{subfigure}[b]{0.3\textwidth}
        \centering
        \includegraphics[width=0.95\linewidth]{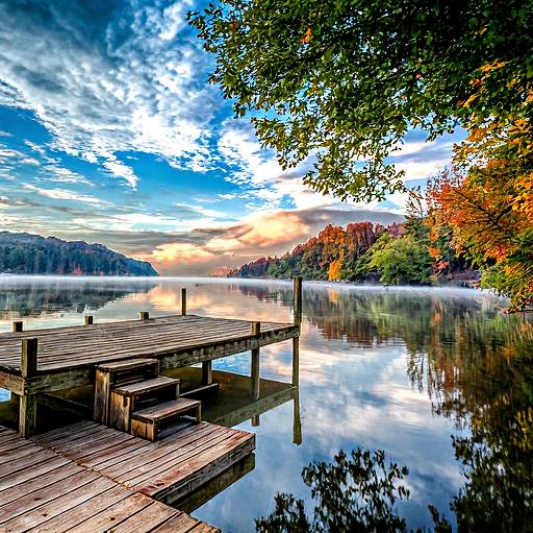}
    \end{subfigure}
    
    % 第二行方法标注
    \vspace{2pt}
    \noindent
    \begin{minipage}[t]{0.3\textwidth}
        \centering SUPIR \cite{yu2024scaling}
    \end{minipage}%
    \begin{minipage}[t]{0.3\textwidth}
        \centering FaithDiff \cite{chen2025faithdiff}
    \end{minipage}%
    \begin{minipage}[t]{0.3\textwidth}
        \centering \textbf{Ours}
    \end{minipage}
    
    \vspace{0.3cm}
    
    % 第三行图片
    \begin{subfigure}[b]{0.3\textwidth}
        \centering
        \includegraphics[width=0.95\linewidth]{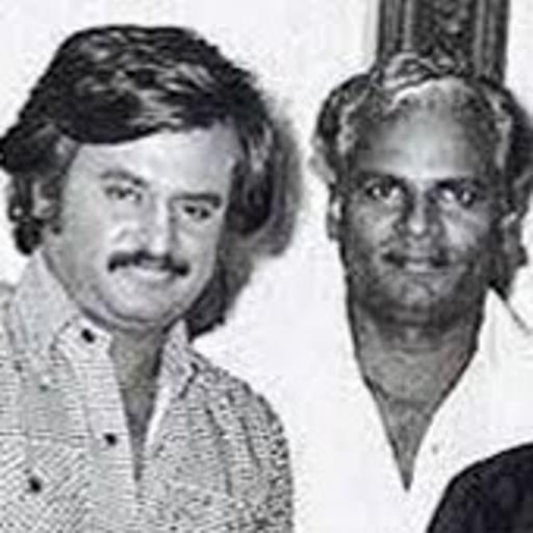}
    \end{subfigure}%
    \begin{subfigure}[b]{0.3\textwidth}
        \centering
        \includegraphics[width=0.95\linewidth]{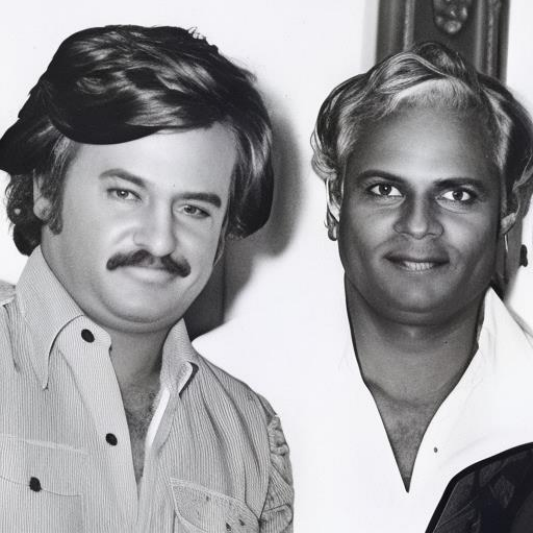}
    \end{subfigure}%
    \begin{subfigure}[b]{0.3\textwidth}
        \centering
        \includegraphics[width=0.95\linewidth]{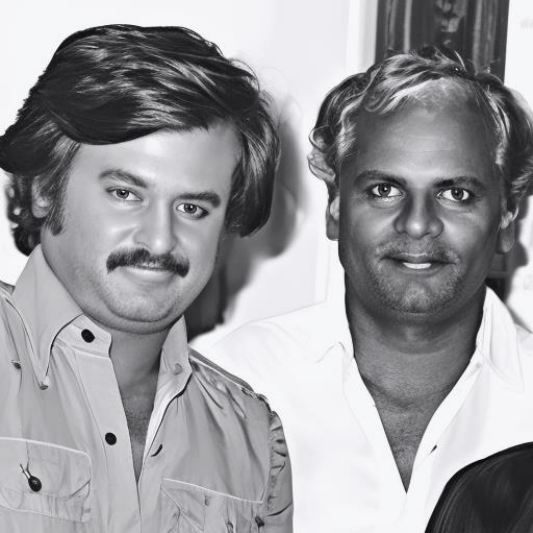}
    \end{subfigure}
    
    % 第三行方法标注
    \vspace{2pt}
    \noindent
    \begin{minipage}[t]{0.3\textwidth}
        \centering LQ Input
    \end{minipage}%
    % 同样减小标注的列间距
    \begin{minipage}[t]{0.3\textwidth}
        \centering DiffBIR \cite{lin2024diffbir}
    \end{minipage}%
    % 同样减小标注的列间距
    \begin{minipage}[t]{0.3\textwidth}
        \centering SeeSR \cite{wu2024seesr}
    \end{minipage}

    % 第四行图片
    \begin{subfigure}[b]{0.3\textwidth}
        \centering
        \includegraphics[width=0.95\linewidth]{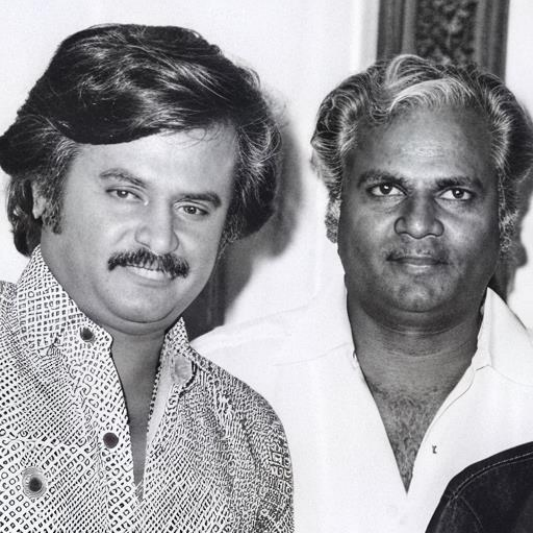}
    \end{subfigure}%
    \begin{subfigure}[b]{0.3\textwidth}
        \centering
        \includegraphics[width=0.95\linewidth]{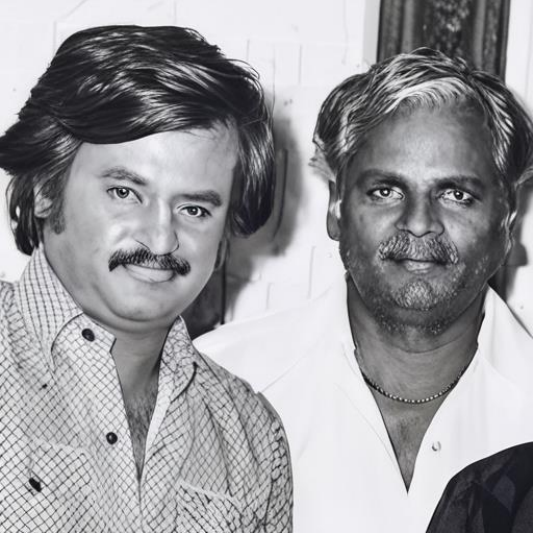}
    \end{subfigure}%
    \begin{subfigure}[b]{0.3\textwidth}
        \centering
        \includegraphics[width=0.95\linewidth]{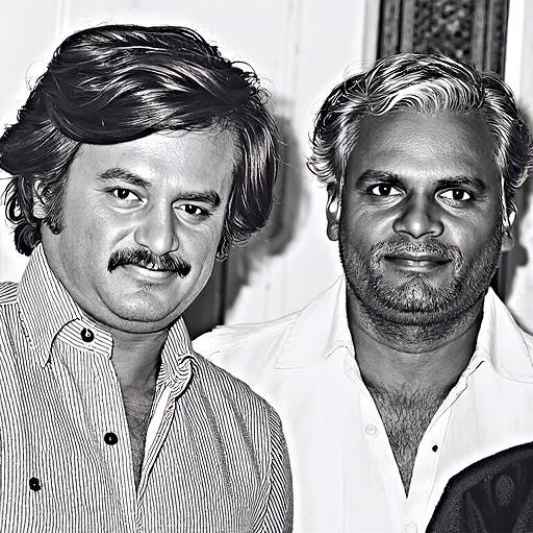}
    \end{subfigure}
    
    % 第四行方法标注
    \vspace{2pt}
    \noindent
    \begin{minipage}[t]{0.3\textwidth}
        \centering SUPIR \cite{yu2024scaling}
    \end{minipage}%
    \begin{minipage}[t]{0.3\textwidth}
        \centering FaithDiff \cite{chen2025faithdiff}
    \end{minipage}%
    \begin{minipage}[t]{0.3\textwidth}
        \centering \textbf{Ours}
    \end{minipage}
    
    \caption{Visual comparisons on real-world benchmarks (2/2). Please zoom in for a better view.}
    \label{fig:real-world-2}
\end{figure*}

\begin{figure*}[ht]
    \centering
    
    % 第一行图片
    \begin{subfigure}[b]{0.23\textwidth}
        \centering
        \includegraphics[width=0.95\linewidth]{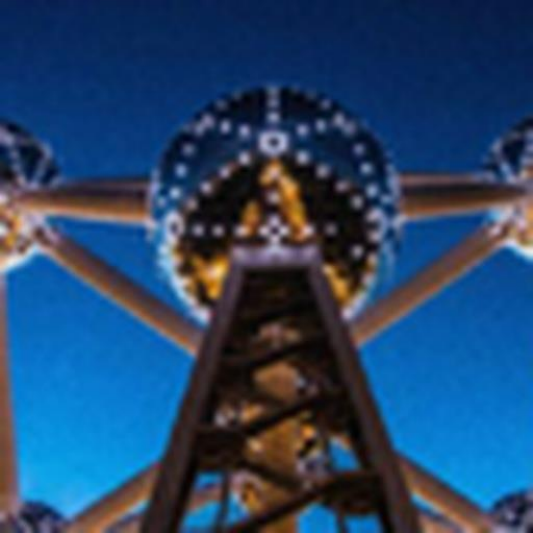}
    \end{subfigure}%
    \begin{subfigure}[b]{0.23\textwidth}
        \centering
        \includegraphics[width=0.95\linewidth]{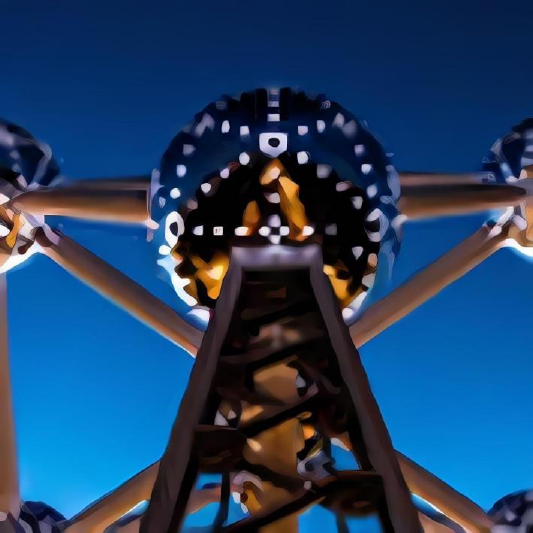}
    \end{subfigure}%
    \begin{subfigure}[b]{0.23\textwidth}
        \centering
        \includegraphics[width=0.95\linewidth]{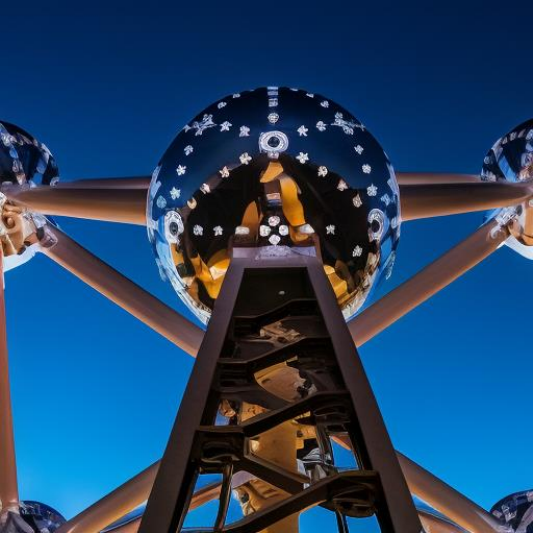}
    \end{subfigure}%
    \begin{subfigure}[b]{0.23\textwidth}
        \centering
        \includegraphics[width=0.95\linewidth]{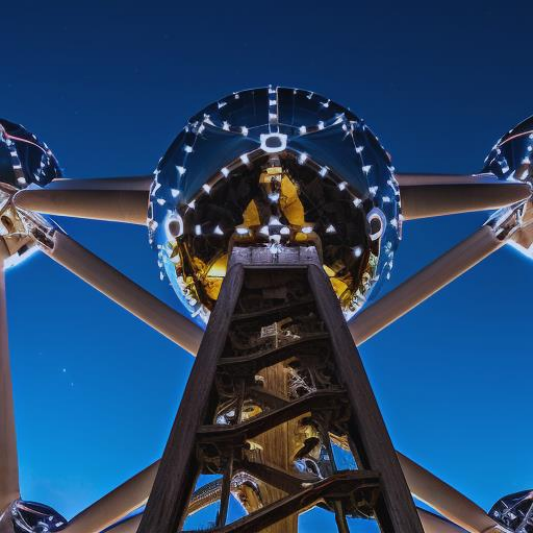}
    \end{subfigure}
    
    % 第一行方法标注
    \vspace{2pt}
    \noindent
    \begin{minipage}[t]{0.23\textwidth}
        \centering LQ Input
    \end{minipage}%
    \begin{minipage}[t]{0.23\textwidth}
        \centering Real-ESRGAN \cite{wang2021real}
    \end{minipage}%
    \begin{minipage}[t]{0.23\textwidth}
        \centering DiffBIR \cite{lin2024diffbir}
    \end{minipage}%
    \begin{minipage}[t]{0.23\textwidth}
        \centering SeeSR \cite{wu2024seesr}
    \end{minipage}%
    
    % 第二行图片
    \begin{subfigure}[b]{0.23\textwidth}
        \centering
        \includegraphics[width=0.95\linewidth]{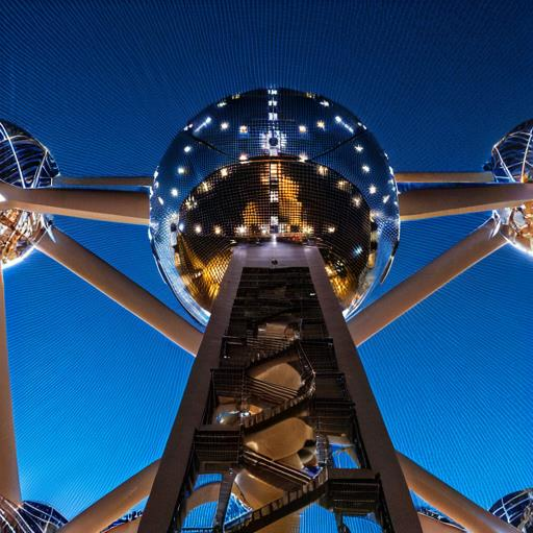}
    \end{subfigure}%
    \begin{subfigure}[b]{0.23\textwidth}
        \centering
        \includegraphics[width=0.95\linewidth]{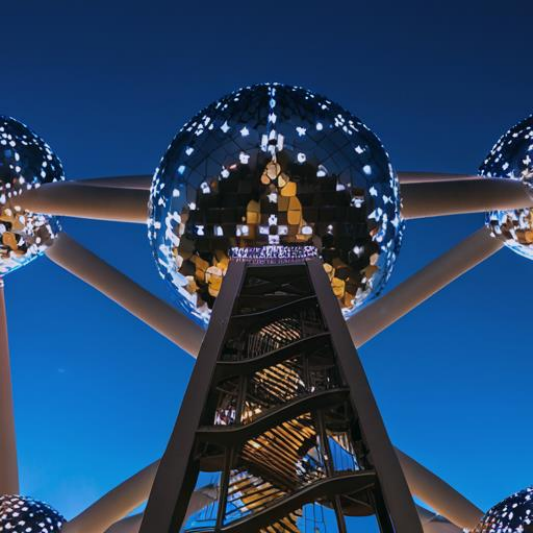}
    \end{subfigure}%
    \begin{subfigure}[b]{0.23\textwidth}
        \centering
        \includegraphics[width=0.95\linewidth]{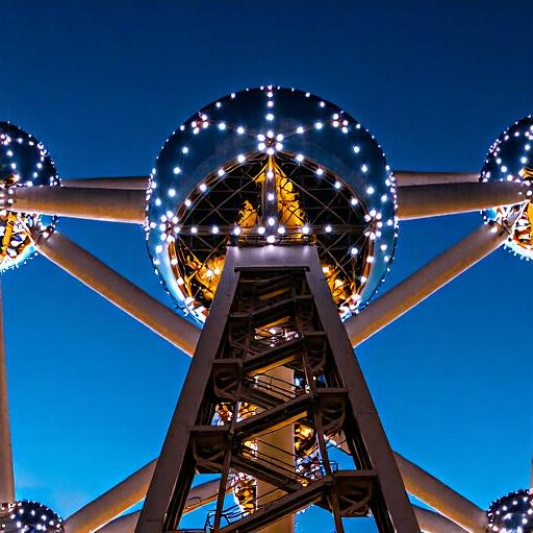}
    \end{subfigure}%
    \begin{subfigure}[b]{0.23\textwidth}
        \centering
        \includegraphics[width=0.95\linewidth]{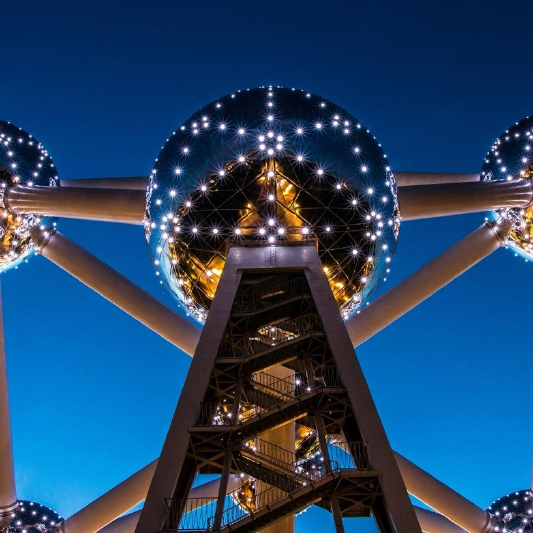}
    \end{subfigure}
    
    % 第二行方法标注
    \vspace{2pt}
    \noindent
    \begin{minipage}[t]{0.23\textwidth}
        \centering SUPIR \cite{yu2024scaling}
    \end{minipage}%
    \begin{minipage}[t]{0.23\textwidth}
        \centering FaithDiff \cite{chen2025faithdiff}
    \end{minipage}%
    \begin{minipage}[t]{0.23\textwidth}
        \centering \textbf{Ours}
    \end{minipage}%
    \begin{minipage}[t]{0.23\textwidth}
        \centering GT
    \end{minipage}
    
    \vspace{0.3cm}
    
    % 第三行图片
    \begin{subfigure}[b]{0.23\textwidth}
        \centering
        \includegraphics[width=0.95\linewidth]{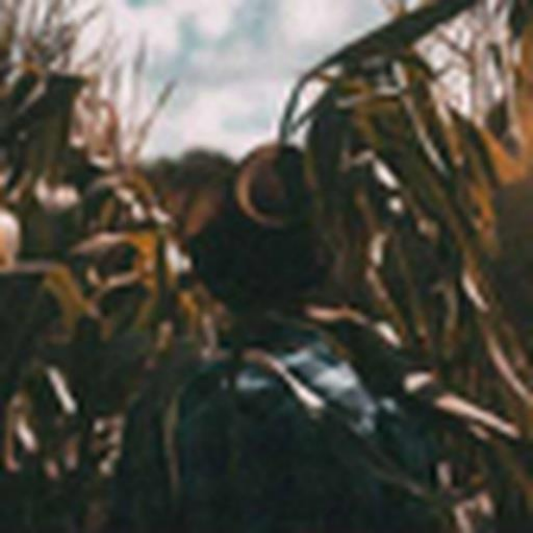}
    \end{subfigure}%
    \begin{subfigure}[b]{0.23\textwidth}
        \centering
        \includegraphics[width=0.95\linewidth]{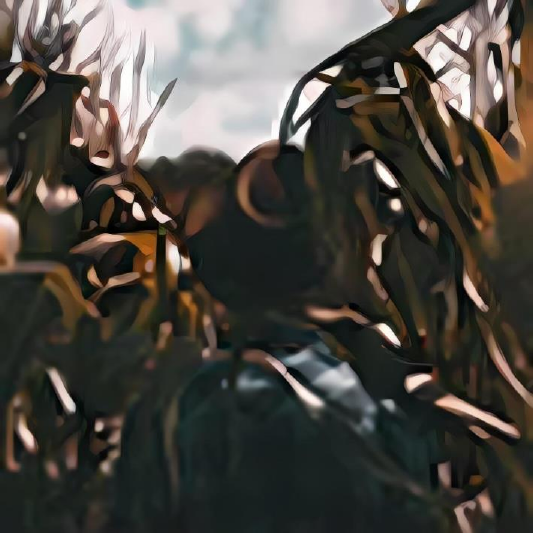}
    \end{subfigure}%
    \begin{subfigure}[b]{0.23\textwidth}
        \centering
        \includegraphics[width=0.95\linewidth]{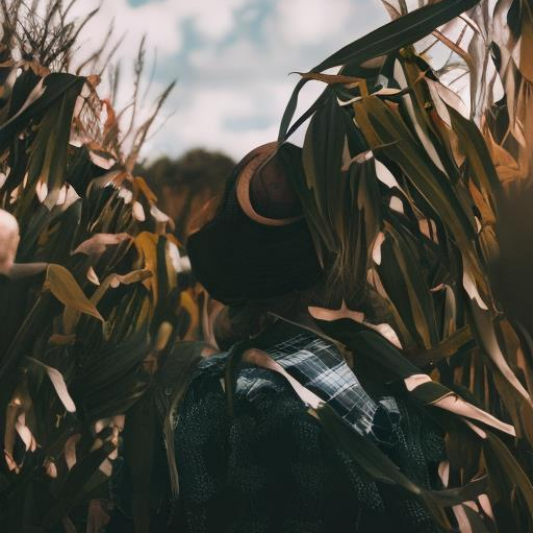}
    \end{subfigure}%
    \begin{subfigure}[b]{0.23\textwidth}
        \centering
        \includegraphics[width=0.95\linewidth]{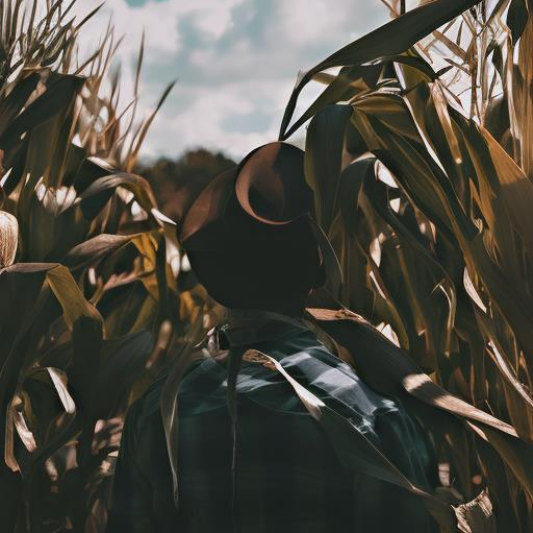}
    \end{subfigure}
    
    % 第三行方法标注
    \vspace{2pt}
    \noindent
    \begin{minipage}[t]{0.23\textwidth}
        \centering LQ Input
    \end{minipage}%
    \begin{minipage}[t]{0.23\textwidth}
        \centering Real-ESRGAN \cite{wang2021real}
    \end{minipage}%
    \begin{minipage}[t]{0.23\textwidth}
        \centering DiffBIR \cite{lin2024diffbir}
    \end{minipage}%
    \begin{minipage}[t]{0.23\textwidth}
        \centering SeeSR \cite{wu2024seesr}
    \end{minipage}%
    
    % 第四行图片
    \begin{subfigure}[b]{0.23\textwidth}
        \centering
        \includegraphics[width=0.95\linewidth]{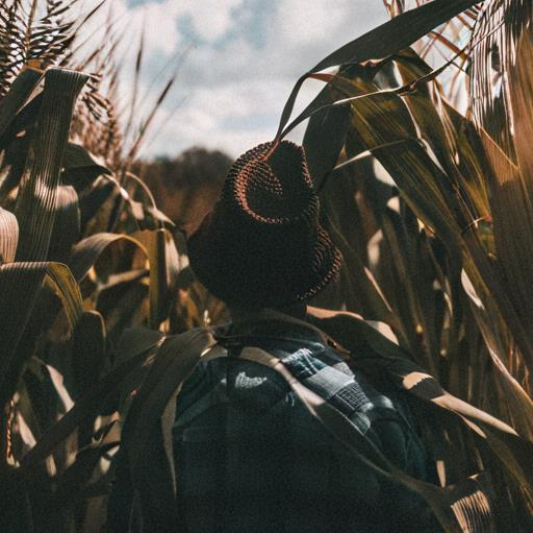}
    \end{subfigure}%
    \begin{subfigure}[b]{0.23\textwidth}
        \centering
        \includegraphics[width=0.95\linewidth]{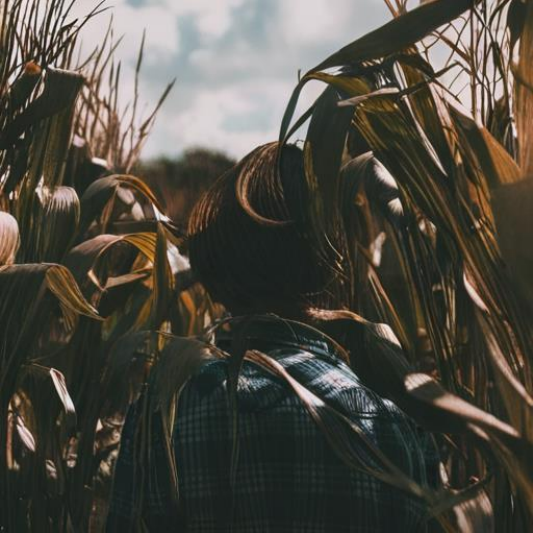}
    \end{subfigure}%
    \begin{subfigure}[b]{0.23\textwidth}
        \centering
        \includegraphics[width=0.95\linewidth]{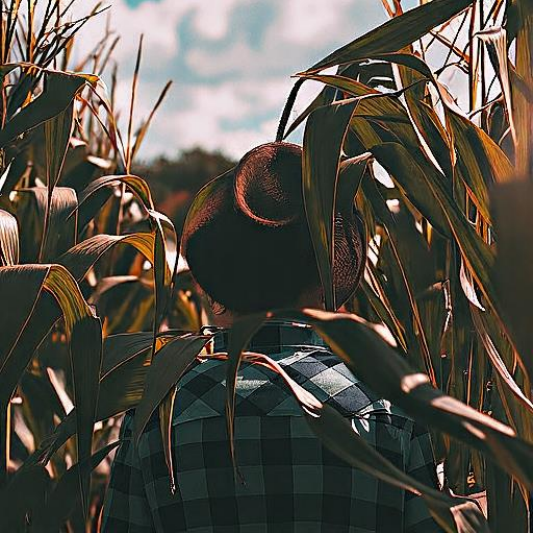}
    \end{subfigure}%
    \begin{subfigure}[b]{0.23\textwidth}
        \centering
        \includegraphics[width=0.95\linewidth]{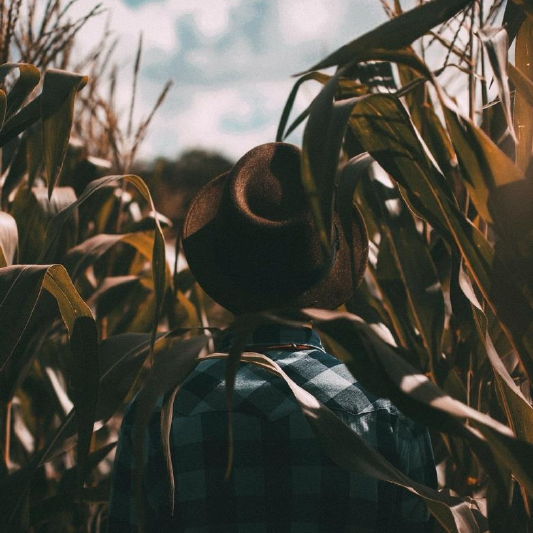}
    \end{subfigure}
    
    % 第四行方法标注
    \vspace{2pt}
    \noindent
    \begin{minipage}[t]{0.23\textwidth}
        \centering SUPIR \cite{yu2024scaling}
    \end{minipage}%
    \begin{minipage}[t]{0.23\textwidth}
        \centering FaithDiff \cite{chen2025faithdiff}
    \end{minipage}%
    \begin{minipage}[t]{0.23\textwidth}
        \centering \textbf{Ours}
    \end{minipage}%
    \begin{minipage}[t]{0.23\textwidth}
        \centering GT
    \end{minipage}
    
    \caption{Visual comparisons on synthetic benchmarks (1/2). Please zoom in for a better view.}
    \label{fig:synthetic-1}
\end{figure*}

\begin{figure*}[ht]
    \centering
    
    % 第一行图片
    \begin{subfigure}[b]{0.23\textwidth}
        \centering
        \includegraphics[width=0.95\linewidth]{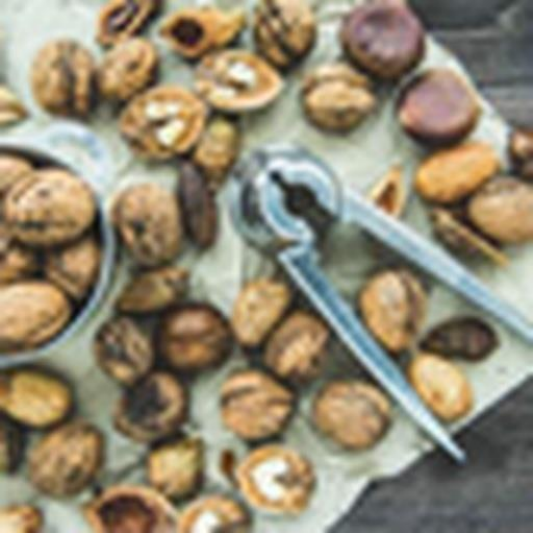}
    \end{subfigure}%
    \begin{subfigure}[b]{0.23\textwidth}
        \centering
        \includegraphics[width=0.95\linewidth]{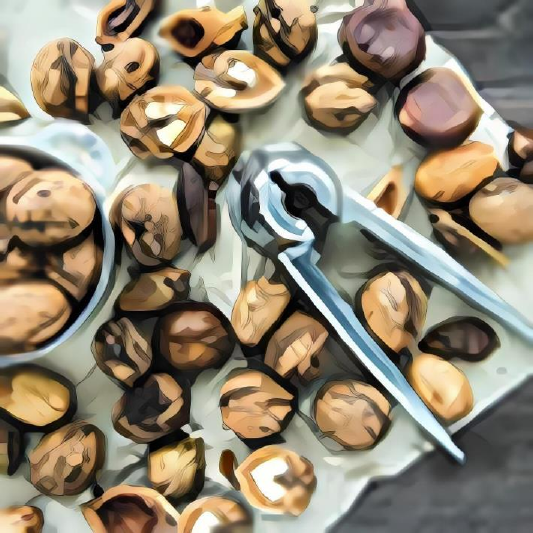}
    \end{subfigure}%
    \begin{subfigure}[b]{0.23\textwidth}
        \centering
        \includegraphics[width=0.95\linewidth]{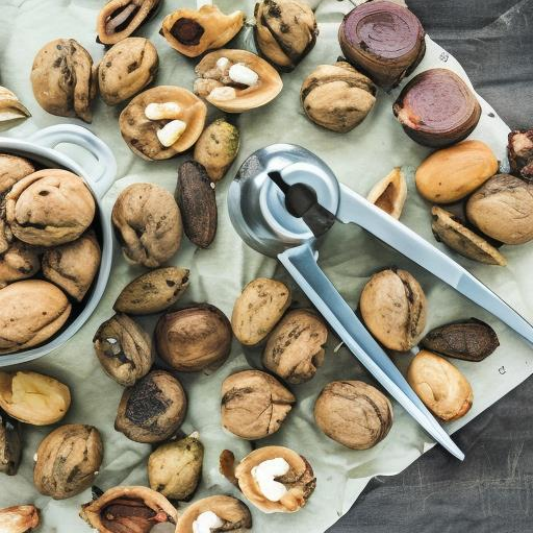}
    \end{subfigure}%
    \begin{subfigure}[b]{0.23\textwidth}
        \centering
        \includegraphics[width=0.95\linewidth]{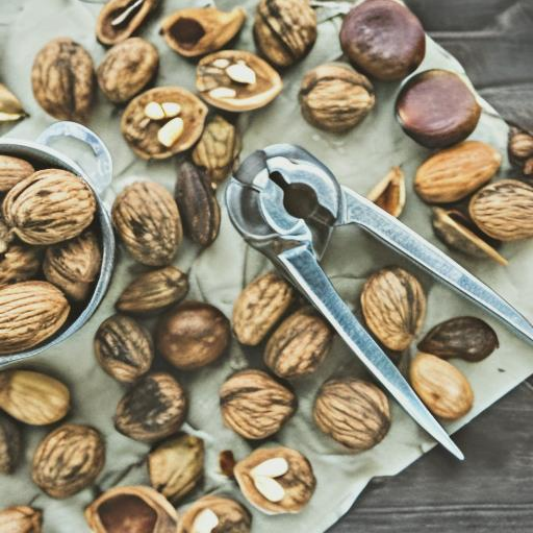}
    \end{subfigure}
    
    % 第一行方法标注
    \vspace{2pt}
    \noindent
    \begin{minipage}[t]{0.23\textwidth}
        \centering LQ Input
    \end{minipage}%
    \begin{minipage}[t]{0.23\textwidth}
        \centering Real-ESRGAN \cite{wang2021real}
    \end{minipage}%
    \begin{minipage}[t]{0.23\textwidth}
        \centering DiffBIR \cite{lin2024diffbir}
    \end{minipage}%
    \begin{minipage}[t]{0.23\textwidth}
        \centering SeeSR \cite{wu2024seesr}
    \end{minipage}%
    
    % 第二行图片
    \begin{subfigure}[b]{0.23\textwidth}
        \centering
        \includegraphics[width=0.95\linewidth]{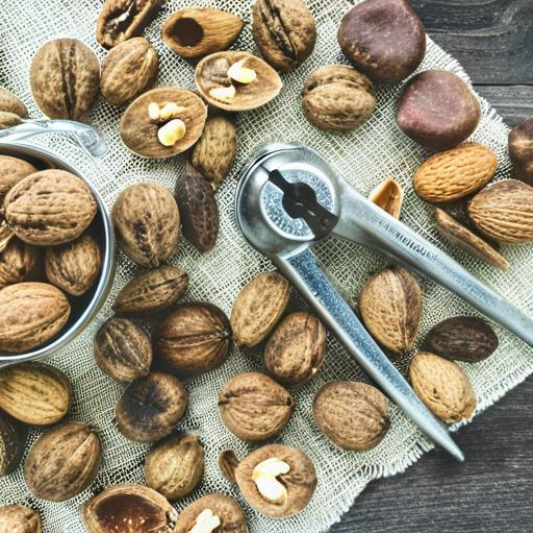}
    \end{subfigure}%
    \begin{subfigure}[b]{0.23\textwidth}
        \centering
        \includegraphics[width=0.95\linewidth]{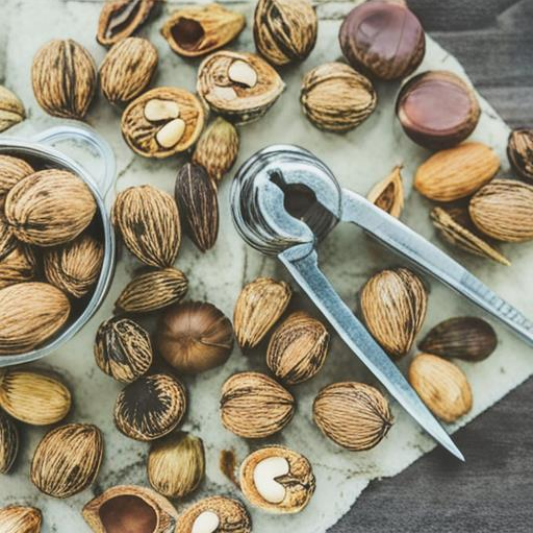}
    \end{subfigure}%
    \begin{subfigure}[b]{0.23\textwidth}
        \centering
        \includegraphics[width=0.95\linewidth]{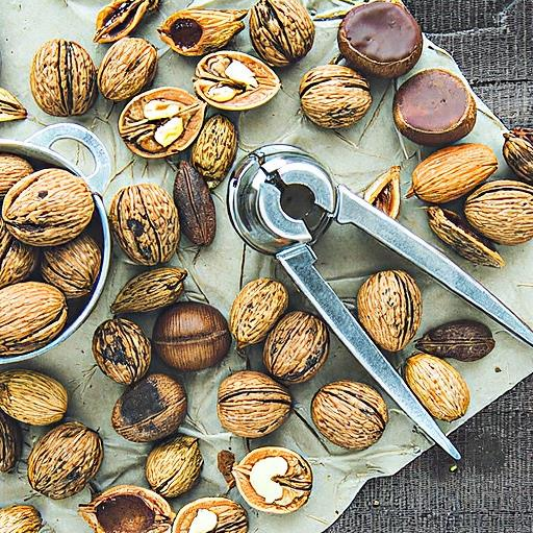}
    \end{subfigure}%
    \begin{subfigure}[b]{0.23\textwidth}
        \centering
        \includegraphics[width=0.95\linewidth]{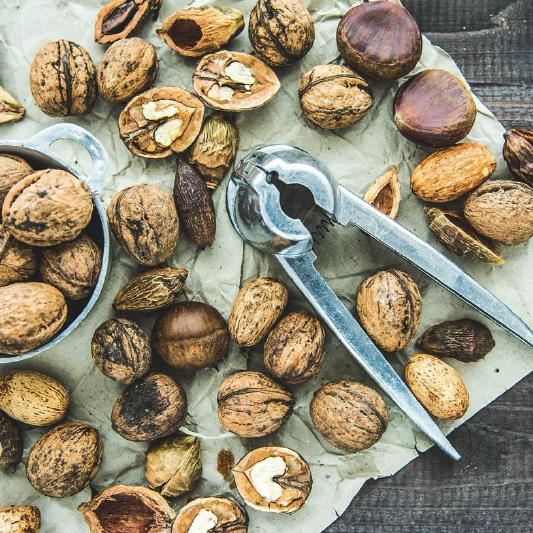}
    \end{subfigure}
    
    % 第二行方法标注
    \vspace{2pt}
    \noindent
    \begin{minipage}[t]{0.23\textwidth}
        \centering SUPIR \cite{yu2024scaling}
    \end{minipage}%
    \begin{minipage}[t]{0.23\textwidth}
        \centering FaithDiff \cite{chen2025faithdiff}
    \end{minipage}%
    \begin{minipage}[t]{0.23\textwidth}
        \centering \textbf{Ours}
    \end{minipage}%
    \begin{minipage}[t]{0.23\textwidth}
        \centering GT
    \end{minipage}
    
    \vspace{0.3cm}
    
    % 第三行图片
    \begin{subfigure}[b]{0.23\textwidth}
        \centering
        \includegraphics[width=0.95\linewidth]{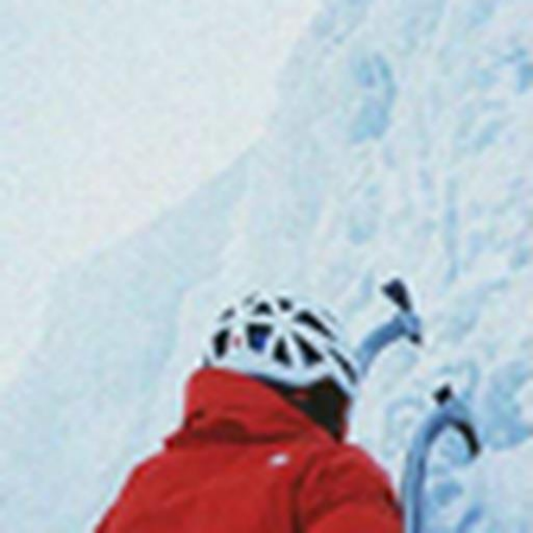}
    \end{subfigure}%
    \begin{subfigure}[b]{0.23\textwidth}
        \centering
        \includegraphics[width=0.95\linewidth]{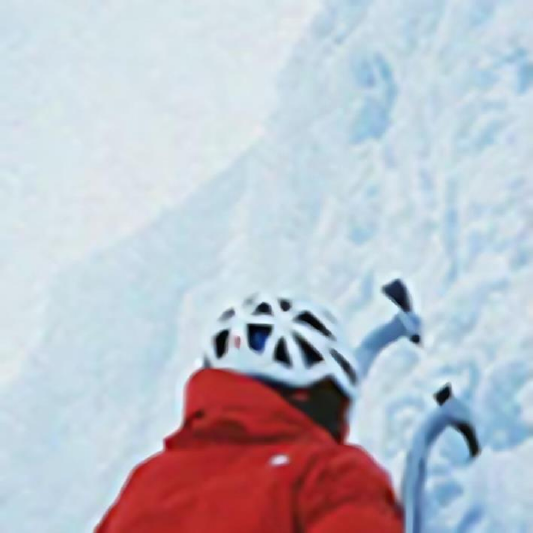}
    \end{subfigure}%
    \begin{subfigure}[b]{0.23\textwidth}
        \centering
        \includegraphics[width=0.95\linewidth]{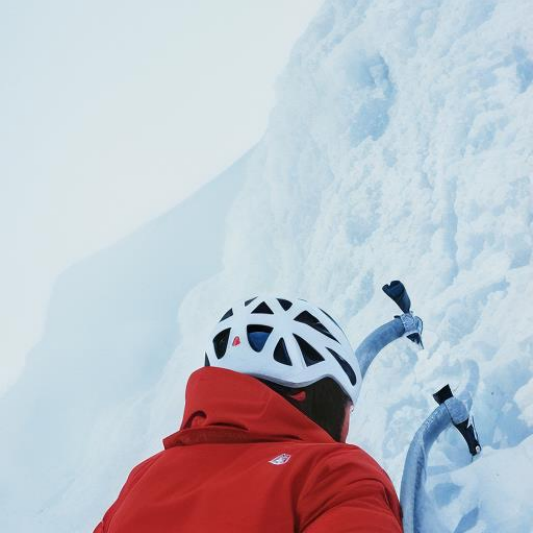}
    \end{subfigure}%
    \begin{subfigure}[b]{0.23\textwidth}
        \centering
        \includegraphics[width=0.95\linewidth]{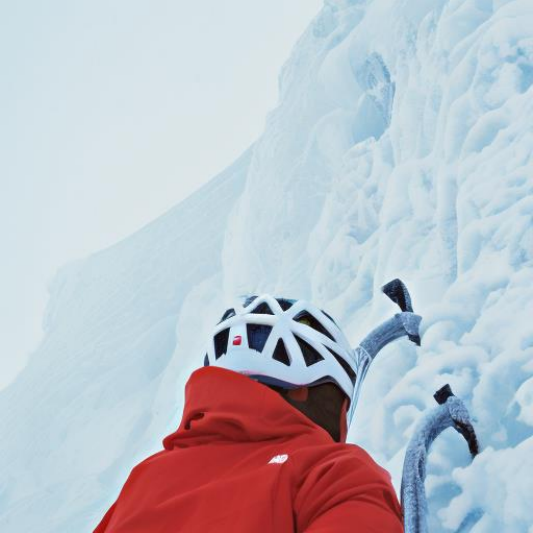}
    \end{subfigure}
    
    % 第三行方法标注
    \vspace{2pt}
    \noindent
    \begin{minipage}[t]{0.23\textwidth}
        \centering LQ Input
    \end{minipage}%
    \begin{minipage}[t]{0.23\textwidth}
        \centering Real-ESRGAN \cite{wang2021real}
    \end{minipage}%
    \begin{minipage}[t]{0.23\textwidth}
        \centering DiffBIR \cite{lin2024diffbir}
    \end{minipage}%
    \begin{minipage}[t]{0.23\textwidth}
        \centering SeeSR \cite{wu2024seesr}
    \end{minipage}%
    
    % 第四行图片
    \begin{subfigure}[b]{0.23\textwidth}
        \centering
        \includegraphics[width=0.95\linewidth]{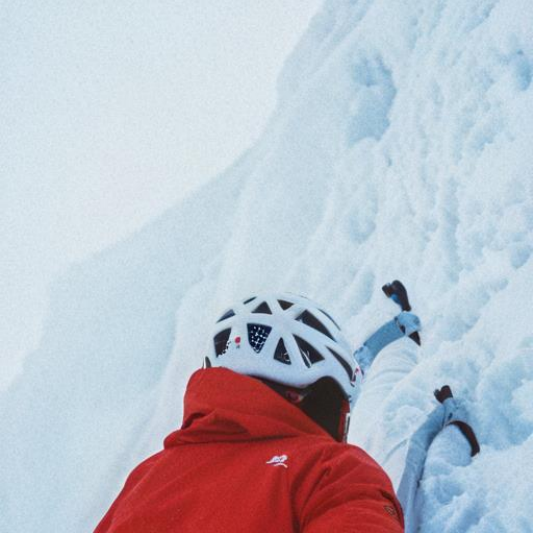}
    \end{subfigure}%
    \begin{subfigure}[b]{0.23\textwidth}
        \centering
        \includegraphics[width=0.95\linewidth]{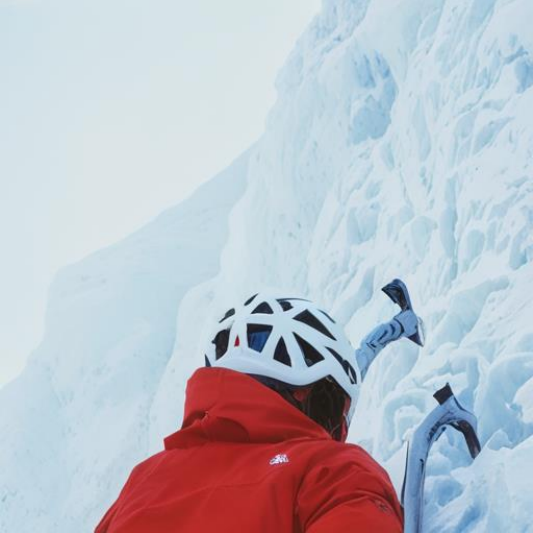}
    \end{subfigure}%
    \begin{subfigure}[b]{0.23\textwidth}
        \centering
        \includegraphics[width=0.95\linewidth]{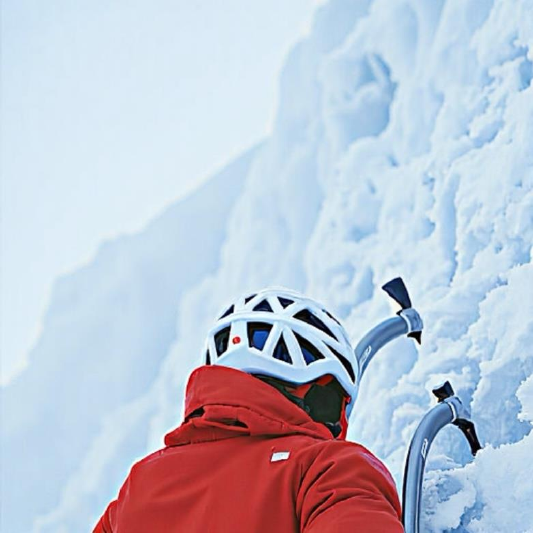}
    \end{subfigure}%
    \begin{subfigure}[b]{0.23\textwidth}
        \centering
        \includegraphics[width=0.95\linewidth]{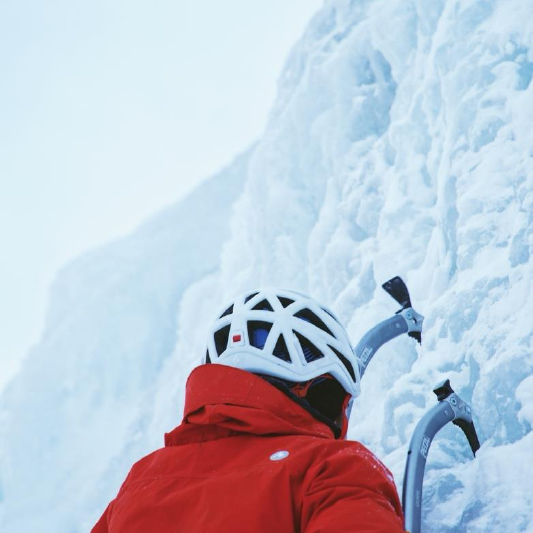}
    \end{subfigure}
    
    % 第四行方法标注
    \vspace{2pt}
    \noindent
    \begin{minipage}[t]{0.23\textwidth}
        \centering SUPIR \cite{yu2024scaling}
    \end{minipage}%
    \begin{minipage}[t]{0.23\textwidth}
        \centering FaithDiff \cite{chen2025faithdiff}
    \end{minipage}%
    \begin{minipage}[t]{0.23\textwidth}
        \centering \textbf{Ours}
    \end{minipage}%
    \begin{minipage}[t]{0.23\textwidth}
        \centering GT
    \end{minipage}
    
    \caption{Visual comparisons on synthetic benchmarks (2/2). Please zoom in for a better view.}
    \label{fig:synthetic-2}
\end{figure*}

We provide more visual comparisons with state-of-the-art image restoration methods on both real-world and synthetic benchmarks in Fig. \ref{fig:real-world-1}, Fig. \ref{fig:real-world-2}, Fig. \ref{fig:synthetic-1} and Fig. \ref{fig:synthetic-2}.

\bibliographystyle{splncs04}
\bibliography{main}
\end{document}